\documentclass[runningheads]{llncs}

\usepackage{accv}
\usepackage{accvabbrv}
\usepackage{graphicx}
\usepackage{booktabs}
\usepackage{placeins}    \AddToHook{package/axessibility/after}{\pdfminorversion=7}
\usepackage[accsupp]{axessibility}
\usepackage{hyperref}
\hypersetup{hidelinks,pdftitle={MonoMIP: Multi-Height Inverse Perspective Mapping for Monocular 3D Object Detection},pdfauthor={Emir Ulurak and Furkan Kirac}}

\graphicspath{{./}}

\newenvironment{widefigure}{\begin{figure}[tb]}{\end{figure}}
\newenvironment{widetable}{\begin{table}[tb]}{\end{table}}

\newcommand{\ours}{MonoMIP}
\newcommand{\R}{\mathbb{R}}
\newcommand{\suppref}[1]{(suppl.\ #1)}

\begin{document}

\title{\texorpdfstring{\ours{}: Multi-Height Inverse Perspective\\Mapping for Monocular 3D Object Detection}{MonoMIP: Multi-Height Inverse Perspective Mapping for Monocular 3D Object Detection}}
\titlerunning{MonoMIP: Multi-Height IPM for Monocular 3D Detection}

\author{Emir Ulurak\inst{1} \and Furkan K{\i}ra\c{c}\inst{1}}
\authorrunning{E. Ulurak and F. K{\i}ra\c{c}}
\institute{Ozyegin University, Istanbul, Turkey\\
\email{emir.ulurak@ozu.edu.tr, furkan.kirac@ozyegin.edu.tr}}

\maketitle

\begin{abstract}
Monocular 3D object detection methods increasingly rely on depth-guided transformers, but they operate entirely in the perspective image plane where spatial relationships are distorted by foreshortening.
We propose \ours{}, a dual-decoder framework that adds complementary 3D geometric reasoning through Bird's-Eye View (BEV) features constructed via \textit{Multi-Height Inverse Perspective Mapping}.
We perform IPM at $K$ elevation planes spanning ground to rooftop, producing an ordered stack of height-conditioned metric BEV features that reduces above-ground misprojection and exposes depth-dependent appearance variation across planes.
A BEV decoder with multi-scale deformable attention aggregates this multi-height evidence per query, while a depth-guided 3D decoder provides complementary perspective-space predictions. Without extra training data, \ours{} attains $25.84\%$ Moderate $\mathrm{AP_{3D|R40}}$ on KITTI \textit{val} and $21.66\%$ on the official \textit{test} set for Car, outperforming recent published monocular methods on both the KITTI and Waymo benchmarks.
\end{abstract}

\section{Introduction}
\label{sec:intro}

Monocular 3D object detection estimates the 3D position, size, and orientation of objects from a single RGB image.
A lone camera is far cheaper and simpler to deploy than the LiDAR~\cite{shi2019pointrcnn,lang2019pointpillars} or stereo~\cite{chen2020dsgn,guo2021ligastereo} rigs used elsewhere, which makes it appealing for autonomous driving and robotics.
The difficulty is geometric: a single perspective image fixes bearing but not range, so infinitely many 3D scenes project to identical pixels and absolute depth cannot be recovered directly.

Depth-guided transformers have advanced monocular 3D detection. MonoDETR~\cite{zhang2023monodetr} uses depth-aware position encodings and decoder attention to incorporate global context. MonoDGP~\cite{pu2025monodgp} decouples 2D and 3D queries and refines depth with geometry-error priors. Both operate in the perspective image plane, where foreshortening distorts spatial relationships.

Bird's-Eye View (BEV) representations provide a metric coordinate system where spatial relationships are preserved regardless of depth.
Multi-camera methods~\cite{philion2020lift,li2022bevformer} have demonstrated the power of BEV for 3D detection, but constructing BEV features from a \emph{single} image remains an open challenge.
Inverse Perspective Mapping (IPM)~\cite{mallot1991ipm} uses only the known camera calibration to map image pixels onto a world plane by pure projective geometry, with no learned parameters and no depth to estimate. This determinism has long made IPM a workhorse of driving perception, spanning obstacle detection, lane detection~\cite{garnett2019lanenet}, and semantic BEV mapping~\cite{reiher2020sim2real}.

\textbf{Key insight.} Multi-height IPM projects image features to several elevation planes instead of committing every feature to the ground plane. For a fixed BEV cell, the planes sample the same image column at different rows, forming an ordered vertical appearance profile: above-ground parts are represented near their correct BEV locations on matching planes, and row offsets follow the $1/Z$ law derived in Section~\ref{sec:multiheight_ipm}. The BEV decoder aggregates this height-conditioned evidence per query; learned features preserve this structure (Figure~\ref{fig:depth_cue_proof}, $r{=}0.75$), and test-time height-order interventions show that the trained level arrangement matters~\suppref{Sec.~E}.

\begin{widefigure}
\centering
\includegraphics[width=\textwidth]{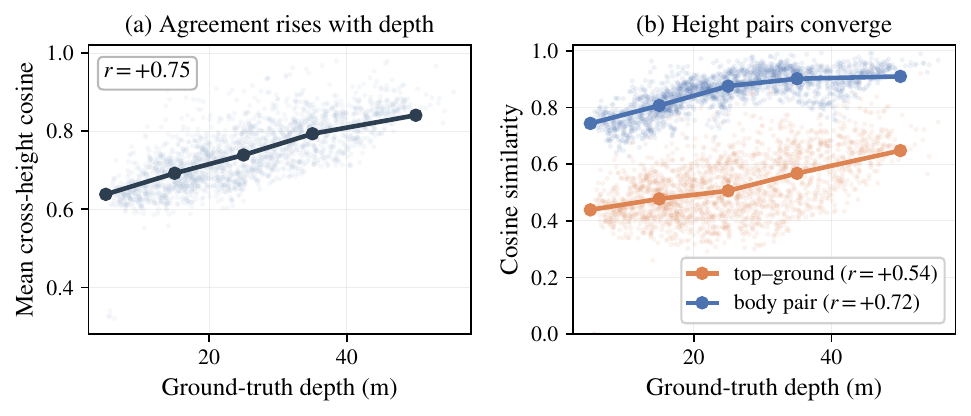}
\caption{\textbf{Cross-height agreement follows the projection geometry.} The row offset between height planes decreases as $1/Z$, so cross-height feature similarity is expected to rise with depth. Learned BEV features preserve this: mean cross-height cosine similarity increases with ground-truth depth ($r{=}0.75$), while wider pairs remain less aligned than adjacent body-level planes.}
\label{fig:depth_cue_proof}
\end{widefigure}

We instantiate this idea in \ours{} (\textbf{Mono}cular 3D Detection via \textbf{M}ulti-Height \textbf{I}nverse \textbf{P}erspective Mapping), a dual-decoder framework combining:
(1) a \textit{3D detection decoder} built on depth-guided deformable transformers~\cite{zhang2023monodetr,pu2025monodgp} that produces object proposals with depth estimates in image space, and
(2) a \textit{BEV decoder} that constructs height-conditioned BEV features via multi-height IPM and refines metric object centers through deformable cross-attention over the stacked BEV features.
Both decoders share a common backbone and are trained jointly end-to-end.

Our main contributions are:
\begin{itemize}
    \item We introduce \textbf{Multi-Height IPM}, a parameter-free, calibration-grounded image-to-BEV transform that projects encoder features onto $K$ fixed elevation planes. Unlike depth-based lifting, its projection does not use predicted per-pixel depth to place BEV features; the detector still predicts object depth.
    \item We show that the resulting \textbf{ordered height stack} is functional: its ordered, height-conditioned structure gives the BEV decoder usable geometric evidence to aggregate per query, improving monocular 3D localization.
    \item \ours{} improves monocular 3D detection, outperforming recent published methods on both KITTI and Waymo, reaching $25.84\%$ Moderate $\mathrm{AP_{3D|R40}}$ on KITTI \textit{val} and $21.66\%$ on the official Car \textit{test} set.
\end{itemize}

\section{Related Work}
\label{sec:related}

\subsection{Monocular 3D Object Detection}

Early monocular 3D detectors predicted boxes from keypoint heatmaps or geometric priors on 2D detections~\cite{zhou2019objects,liu2020smoke,brazil2019m3drpn} and modeled depth confidence through geometry uncertainty~\cite{lu2021gupnet}. Pseudo-LiDAR~\cite{wang2019pseudolidar} converts estimated depth to point clouds, whereas DD3D~\cite{park2021pseudolidar} uses depth pretraining without this intermediate representation. Others handle truncated objects~\cite{zhang2021monoflex}, analyze localization error~\cite{ma2021monodle}, or fuse probabilistic and geometric depth cues~\cite{wang2021fcos3d,wang2022pgd,shi2021geometry}.

Transformer-based detectors~\cite{carion2020detr,zhu2021deformable} include MonoDETR~\cite{zhang2023monodetr}, which predicts foreground depth and feeds it to the decoder through depth-aware positional encodings. MonoDGP~\cite{pu2025monodgp} separates 2D and 3D query streams and refines depth using geometry-error priors. Others target occlusion~\cite{jiang2024monomae}, learnable sample selection~\cite{li2024monolss}, diffusion-based box generation~\cite{ranasinghe2024monodiff}, top-down denoising~\cite{liu2023monoxiver}, and global depth encoding~\cite{huang2022monodtr}. More recently, MonoCoP~\cite{zhang2026monocop} models inter-attribute correlations with an uncertainty-guided selector on the MonoDGP framework, while MonoIA~\cite{zhang2026monoia} uses language and foundation-model intrinsic conditioning for unseen focal lengths and multi-dataset training. Our approach adds a BEV decoder over geometry-based IPM features, providing metric spatial reasoning difficult to achieve in image space alone.

\subsection{Bird's-Eye View Representation}

Constructing BEV features from images has been approached through depth-based lifting~\cite{philion2020lift,huang2021bevdet,li2023bevdepth,li2022bevstereo}, which predicts per-pixel depth distributions to ``lift'' image features into 3D then collapses them onto BEV; attention-based projection with BEV queries~\cite{li2022bevformer}; and geometric projection~\cite{roddick2019oft,mallot1991ipm}, which uses known camera geometry for deterministic image-to-BEV mapping. PETR~\cite{liu2022petr} instead uses object queries over 3D position-aware image features. CaDDN~\cite{reading2021caddn} combined categorical depth distributions with BEV projection for monocular detection.
Our multi-height IPM belongs to the geometric-projection family but extends classical IPM~\cite{mallot1991ipm}, which assumes a single flat ground plane (exact only for road-level content, as used in lane and free-space estimation~\cite{reiher2020sim2real,yang2021projecting}), from one plane to multiple elevations. Rather than learning residual corrections (which entangle geometry with learned network capacity), we perform \emph{several geometry-defined projections} and let the BEV decoder aggregate the appropriate height-conditioned evidence per query, keeping the mapping interpretable. SHOT~\cite{song2021shot} also uses stacked height homographies for multi-view pedestrian occupancy. Our contribution couples single-view ordered reference contrasts with proposal-conditioned metric BEV refinement, rather than projection or deformable attention alone.

\section{Method}
\label{sec:method}

\subsection{Motivation and Overview}

\begin{figure}[!t]
\centering
\includegraphics[width=0.96\textwidth]{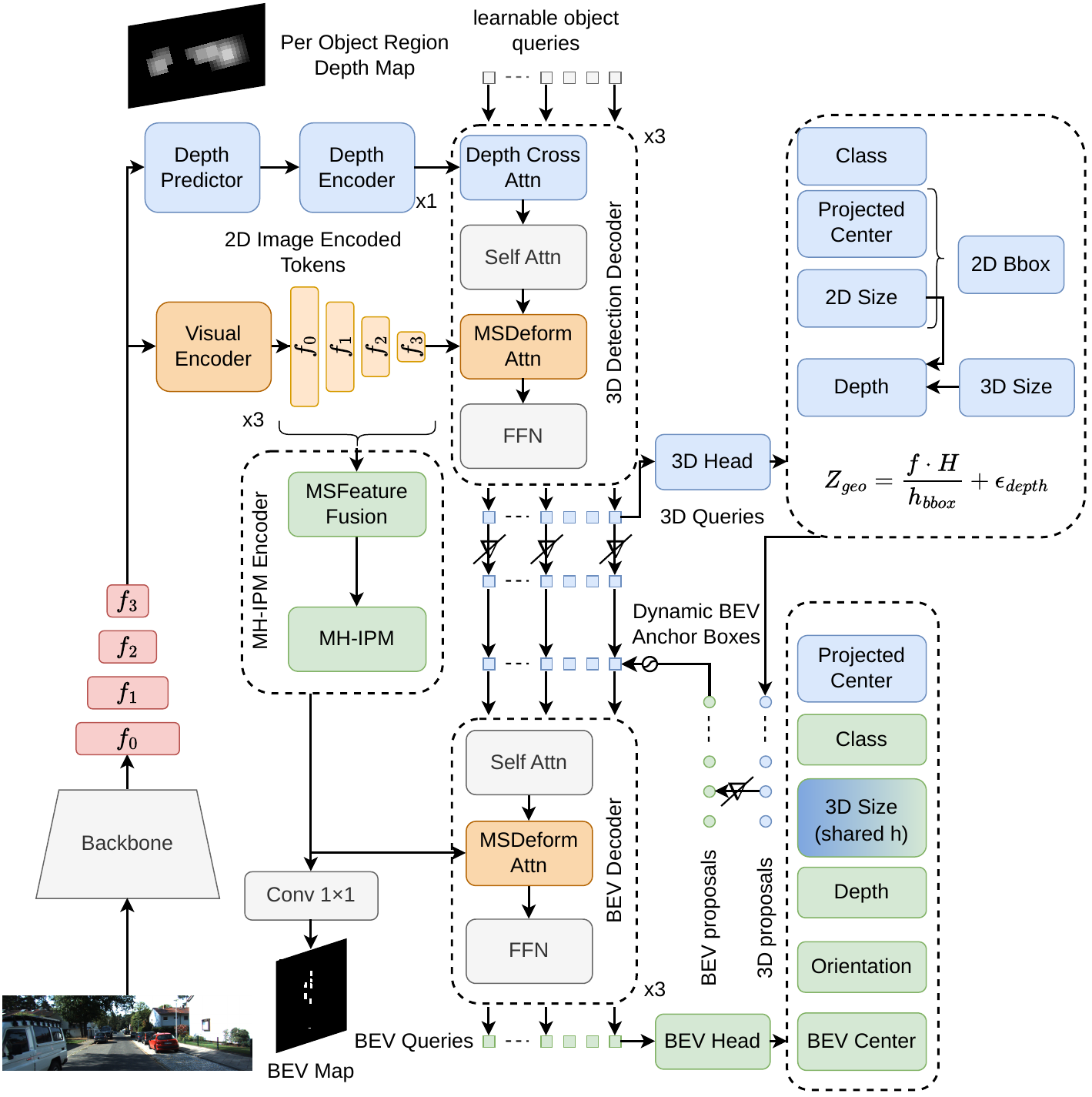}
\caption{\textbf{Overall architecture of \ours{}.} A shared backbone extracts multi-scale features. The image-space decoder produces depth-aware object proposals, while the BEV decoder builds multi-height IPM features and refines metric boxes through deformable attention across height planes. Both decoders share queries and are trained jointly. The schematic's depth expression includes the learned residual and denotes $\hat z$.}
\label{fig:overview}
\end{figure}

Given a monocular image $\mathbf{I} \in \R^{3 \times H_I \times W_I}$, camera intrinsics $\mathbf{K} \in \R^{3 \times 3}$, and a calibrated camera height $H_\mathrm{cam}$ relative to the ground plane, \ours{} estimates 3D bounding boxes parameterized by center $(x, y, z)$, dimensions $(h, w, l)$, and heading angle $\theta$.

Figure~\ref{fig:overview} shows the shared backbone and the image-space and metric BEV decoders, connected by multi-height IPM.

\subsubsection{Why Inverse Perspective Mapping.}
IPM offers a direct, learning-free path from image to BEV: given the camera calibration, it maps each pixel onto a world plane by pure projective geometry, predicting no depth and adding no parameters. The prevailing alternative is depth-based lifting, such as Lift-Splat-Shoot (LSS)~\cite{philion2020lift} and related depth-then-splat methods~\cite{huang2021bevdet,li2023bevdepth,reading2021caddn}, which predict a per-pixel depth distribution, lift features into a camera frustum, and splat them onto the BEV grid. With multi-view or LiDAR-supervised depth this can be effective, but monocular metric depth remains ill-posed: a single image does not uniquely determine range, and uncertain pixel-depth estimates can spread evidence across incorrect BEV cells. MH-IPM instead places features using known calibration and assumed height planes, avoiding predicted pixel depth in the view transform. This does not resolve monocular depth ambiguity; the detector still predicts object depth. A single ground plane misprojects above-ground content, which motivates the multiple-height projections below.

\subsection{Multi-Height Inverse Perspective Mapping}
\label{sec:multiheight_ipm}

\subsubsection{Multi-Scale Feature Fusion.}
\begin{figure}[!t]
\centering
\includegraphics[width=\textwidth]{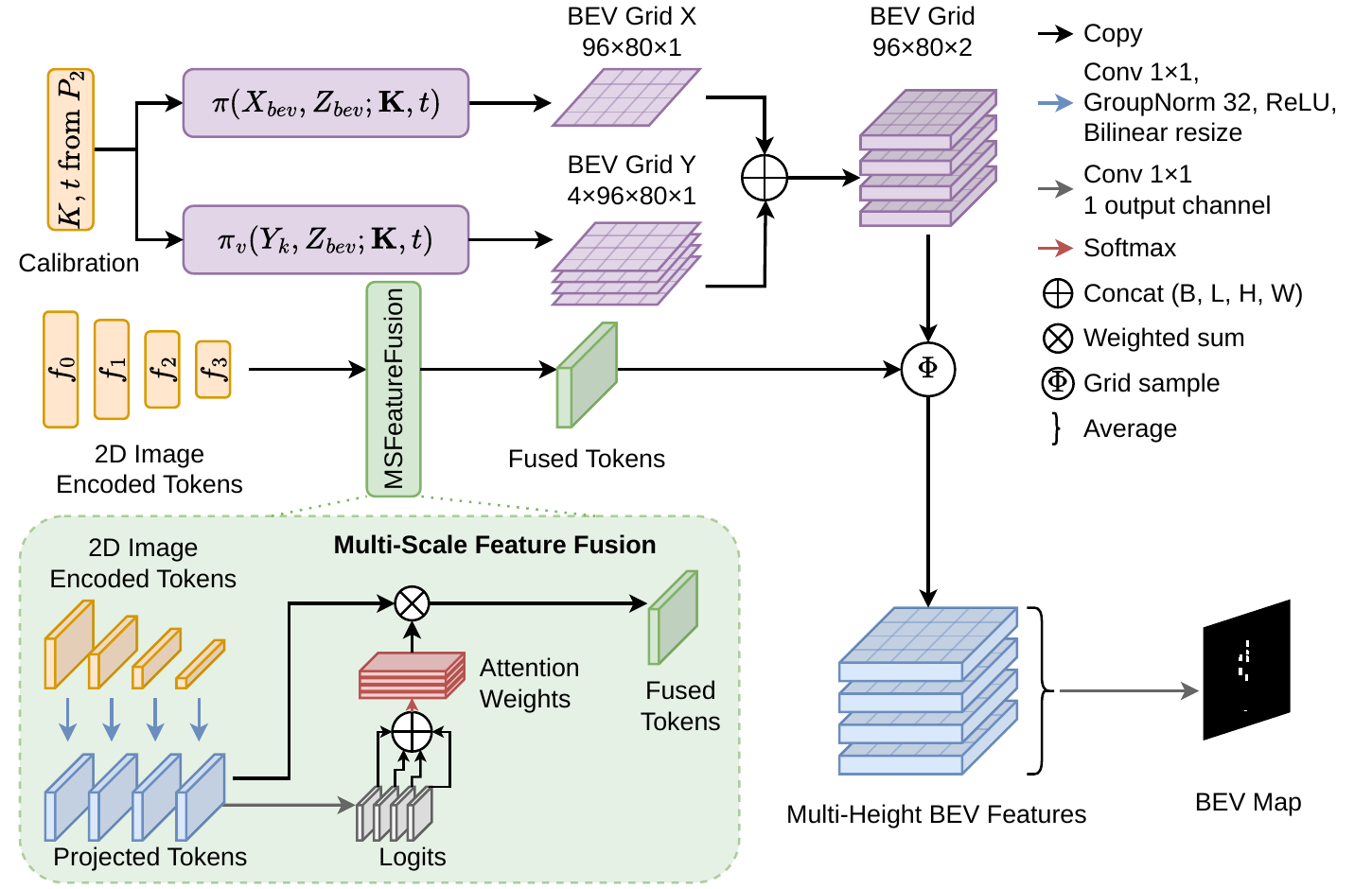}
\caption{\textbf{Multi-Height IPM pipeline.} \textbf{Bottom-left:} Multi-scale encoded features ($f_0$--$f_3$) are fused via learned attention into a single feature map ($F_\mathrm{fused}$). \textbf{Top:} For each of $K{=}4$ height planes, the rectified KITTI projection matrix $\mathbf{P}_2$ provides intrinsic parameters $\mathbf{K}$ and small offsets $t_x, t_y$; together with the fixed height planes $Y_k$, they define a mapping $\pi(X, Z, Y_k; \mathbf{K}, t)$ from each BEV grid cell to an image coordinate $(u, v_k)$. Features are sampled via \texttt{grid\_sample} (bilinear interpolation), producing $K$ BEV feature maps. \textbf{Right:} Reference contrasts (Eq.~\eqref{eq:cross_height}, omitted from the schematic) are normalized via GroupNorm and stacked as multi-scale inputs to the BEV decoder. The IPM sampling grid is geometry-based with no learned parameters; only the feature fusion and normalization layers are trainable.}
\label{fig:mhipm_pipeline}
\end{figure}

Before IPM, we fuse the image encoder's multi-scale tokens through learned attention (Figure~\ref{fig:mhipm_pipeline}). Encoded tokens carry the deformable encoder's global context rather than only local backbone features. Each level is projected to $d{=}256$ channels by $1{\times}1$ convolution, GroupNorm and ReLU, then bilinearly resized to the level-2 resolution. A separate $1{\times}1$ convolution predicts per-level, single-channel logits, which are concatenated and softmax-normalized across levels:
\begin{equation}
    \mathbf{F}_\mathrm{fused} = \sum_{l=1}^{L} \alpha_l \cdot \phi_l(\mathbf{F}_l), \quad \alpha_l = \frac{\exp(z_l)}{\sum_{l'}\exp(z_{l'})}, \quad z_l = \psi(\phi_l(\mathbf{F}_l)),
    \label{eq:fusion}
\end{equation}
where $\phi_l$ is a $1{\times}1$ conv projection and $\psi$ produces per-pixel attention logits. The fused map adaptively draws from high-resolution features (nearby objects) or semantic features (distant objects) and feeds all $K$ height-level IPM projections.

\subsubsection{Multi-Height Projection.}
Classical IPM projects image features onto a single ground plane via known camera parameters. This is exact for ground-level content, but above-ground objects are warped to incorrect BEV locations. With the camera $Y$-axis pointing downward and the ground plane at $Y{=}H_\mathrm{cam}$, a point's height above the ground is $h{=}H_\mathrm{cam}-Y$, and treating it as if it lay \emph{on} the ground mislocates its depth by
\begin{equation}
    \Delta Z = Z \cdot \frac{h}{H_\mathrm{cam} - h},
    \label{eq:ipm_error}
\end{equation}
where $Z$ is the true depth. The error vanishes on the ground ($h{=}0$) and grows with both height and depth. For a KITTI camera ($H_\mathrm{cam} \approx 1.65$\,m), a point $0.5$\,m above the ground at $Z = 30$\,m is displaced by $\Delta Z \approx 13$\,m, diverging as a point nears camera height (e.g.\ a car roof, $h \approx 1.40$\,m). \ours{} does not read this off as a box error: each BEV query is seeded by the image-branch depth and refined (Section~\ref{sec:bev_decoder}), so a single plane stays workable but leaves height and depth entangled, forcing the decoder to infer from the image seed the vertical cue that multi-height IPM supplies directly.

To fix this, we perform IPM at $K$ distinct height levels $\{Y_k\}_{k=1}^{K}$, each a horizontal plane at elevation $Y_k$ in camera coordinates. For a BEV grid cell $(X_i, Z_j)$ at height $Y_k$, the corresponding image coordinate is
\begin{equation}
    u = f_x \frac{X_i - t_x}{Z_j} + c_x, \qquad
    v = f_y \frac{Y_k - t_y}{Z_j} + c_y,
    \label{eq:ipm_projection}
\end{equation}
where $f_x, f_y, c_x, c_y$ are the rectified intrinsic parameters from the KITTI camera-2 projection matrix $\mathbf{P}_2$, and $t_x{=}{-}\mathbf{P}_2[0,3]/f_x$, $t_y{=}{-}\mathbf{P}_2[1,3]/f_y$ are the small rectified camera offsets. Features are sampled by bilinear interpolation from the fused image feature map, and pixels projecting outside the image boundary are masked to zero~\suppref{Sec.~B}.

For scene points lying on a selected height plane $Y_k$, the corresponding IPM map places their features at the correct BEV location under the calibrated flat-ground camera model. The resulting stack of $K$ BEV feature maps $\{\mathbf{B}_k \in \R^{d \times U_z \times U_x}\}_{k=1}^K$ is a discrete volumetric representation, each level a different vertical slice of the scene.

\subsubsection{Ordered Cross-Height Evidence.}
With calibration fixed, Eq.~\eqref{eq:ipm_projection} makes the image column $u$ independent of the selected height plane, while the row $v$ varies with $Y_k$; therefore, for a fixed BEV cell, all $K$ planes sample the \textit{same image column} but \textit{different rows}. The vertical shift between two height planes $Y_k$ and $Y_l$ is:
\begin{equation}
    \Delta v_{kl} = f_y \cdot \frac{Y_k - Y_l}{Z_j}.
    \label{eq:vertical_shift}
\end{equation}
The offset $t_y$ cancels in this difference, so the cross-height shift keeps the same $1/Z$ form (Figure~\ref{fig:multiheight_ipm}). The height stack should therefore be viewed as an ordered set of depth-dependent appearance samples: near objects produce more distinct samples across planes, while far objects produce more similar samples. This structure is not an explicit depth estimator; it gives the BEV decoder a calibrated vertical evidence profile per metric cell.

\begin{widefigure}
\centering
\includegraphics[width=\textwidth]{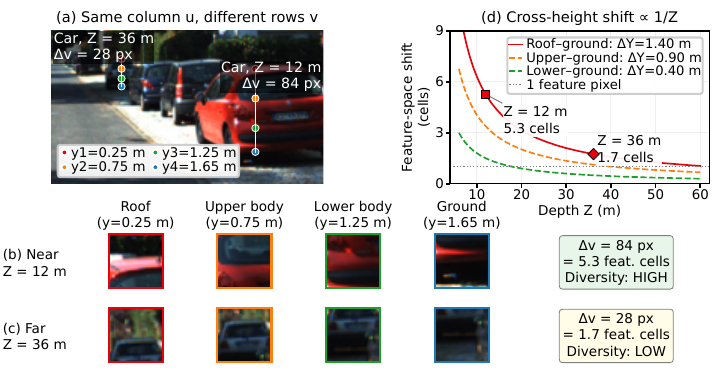}
\caption{\textbf{Cross-height geometry on a KITTI scene.} The four height planes sample the same image column at depth-dependent rows. Near objects produce distinct roof-to-ground samples, while far objects produce more similar samples. The row shift follows the $1/Z$ law, matching Eq.~\eqref{eq:vertical_shift}.}
\label{fig:multiheight_ipm}
\end{widefigure}

\subsubsection{Reference-Anchored Representation.}
We do not feed the $K$ projected maps to the BEV decoder as raw absolute features. IPM is geometrically exact for points on its assumed plane. We use the largest-$Y$ plane as the reference, coinciding with ground in the default configuration, and represent the remaining planes as differences from it:
\begin{equation}
    \mathbf{B}'_0 = \mathbf{B}_\mathrm{ref}, \qquad
    \mathbf{B}'_k = \mathbf{B}_k - \mathbf{B}_\mathrm{ref}, \;\; k\in\{1,2,3\},
    \label{eq:cross_height}
\end{equation}
where $Y_1<Y_2<Y_3<Y_4$ and $\mathbf{B}_\mathrm{ref}=\mathbf{B}_4$. Each difference separates evidence at another height from the reference feature. Before per-level normalization, this linear re-parameterization is information-preserving ($\mathbf{B}_k=\mathbf{B}'_k+\mathbf{B}'_0$). We feed levels in the order $(\mathbf{B}'_0,\mathbf{B}'_3,\mathbf{B}'_2,\mathbf{B}'_1)$, from the reference to increasing vertical baseline. The trained decoder is sensitive to test-time reordering~\suppref{Sec.~E}. Trained coverage variants retain $K{=}4$ without an exact ground plane, showing that the reference need not coincide with physical ground. Figure~\ref{fig:depth_cue_proof} further shows that learned features preserve depth-dependent cross-height agreement.

\subsubsection{Height Plane Selection.}
We place the $K{=}4$ planes at near-uniform $0.5$\,m intervals, spanning the ground plane $Y_4{=}H_\mathrm{cam}{=}1.65$\,m up to the upper plane $Y_1{=}0.25$\,m in the rectified camera frame. The graded spacing lets each level contribute distinct height-conditioned evidence, so the decoder can combine samples aligned with different object parts rather than near-duplicate ground-plane projections. Uniform spacing needs no per-part tuning; only the overall range is scene-dependent, and $K$ is validated by ablation (Table~\ref{tab:ablations}a).

\subsubsection{BEV Occupancy Map.}
We add a lightweight BEV occupancy head that averages the $K$ height features and predicts a single-channel confidence map, supervised by BCE + Dice loss against rasterized ground-truth occupancy. It is an \emph{auxiliary read-out} that shapes the shared BEV features during training; it is not fed back into the decoder and is unused at inference.

\subsection{3D Detection Decoder}
\label{sec:det2d}

Following MonoDETR~\cite{zhang2023monodetr}, the image branch is a depth-guided decoder. The multi-scale image features carry sinusoidal position encodings, and a depth predictor produces a per-pixel foreground depth distribution over $N_d$ LID bins~\cite{reading2021caddn}. An $N_\mathrm{enc}$-layer multi-scale deformable encoder~\cite{zhu2021deformable} enriches the features, and each of the $N_\mathrm{dec}$ decoder layers applies depth cross-attention to the dense depth features, query self-attention, multi-scale deformable cross-attention to the image features, and a feed-forward network. Prediction heads output class, 2D box, 3D dimensions $(h,w,l)$ as learned absolute sizes, and per-object depth. This decoder does not predict heading. The 12-bin/residual angle head belongs only to the BEV decoder. Following the geometry-error prior of MonoDGP~\cite{pu2025monodgp}, the depth head does not regress $\hat{z}$ directly: it forms a geometric term $f_x\,h_\mathrm{3D}/h_\mathrm{2D}$ from the predicted physical height $h_\mathrm{3D}$, the 2D box pixel height $h_\mathrm{2D}$, and focal length $f_x$, and adds a learned residual $\Delta z$, so $\hat{z}{=}f_x h_\mathrm{3D}/h_\mathrm{2D}+\Delta z$ with an aleatoric log-scale $\log\sigma_z$~\cite{kendall2017uncertainties}.

\subsection{Dense Depth Prediction}
\label{sec:depth_predictor}

A dense depth prediction module fuses multi-scale backbone features through a lightweight convolutional encoder and predicts per-pixel depth distributions over $N_d + 1$ bins (including background). The dense depth features serve as context for the decoder's depth cross-attention. Beyond depth awareness, this attention anchors image-space queries on object regions, balancing 2D localization with 3D attribute regression.

\subsubsection{Gaussian 3D-Center Depth Targets.}
MonoDETR~\cite{zhang2023monodetr} fills 2D boxes with object-depth labels, without spatially emphasizing the projected 3D center.
We construct the depth target using a spatial Gaussian centered at the projected 3D object center $(\tilde{c}_x, \tilde{c}_y)$:
\begin{equation}
    G(u, v) = \exp\!\Bigl(-\frac{(u - \tilde{c}_x)^2}{2\sigma_x^2} - \frac{(v - \tilde{c}_y)^2}{2\sigma_y^2}\Bigr),
    \label{eq:gaussian_depth}
\end{equation}
where $\sigma_x = \max(\sigma_s \cdot w_\mathrm{box}, \sigma_\mathrm{min})$ and $\sigma_y = \max(\sigma_s \cdot h_\mathrm{box}, \sigma_\mathrm{min})$ are proportional to box dimensions. Within each object's box, the metric target is $d_\mathrm{target}(u,v)=d_iG(u,v)+d_\mathrm{max}[1-G(u,v)]$, where $d_i$ is its ground-truth center depth and $d_\mathrm{max}=60$\,m is the background depth. A small $p {\times} p$ patch at the 3D center is anchored to exact ground-truth depth. Overlapping objects are combined by per-cell minimum depth. The depth loss is KL divergence between predicted and target distributions, weighted by foreground emphasis ($13\times$).

\subsection{BEV Decoder}
\label{sec:bev_decoder}

The BEV decoder localizes objects in the metric BEV coordinate system using the multi-height IPM features.

\subsubsection{2D-to-BEV Projection.}
For each proposal from the 3D detection decoder, we unproject the predicted image center $(u, v)$ and depth $\hat{z}$ into BEV coordinates:
\begin{equation}
    X = \frac{(u - c_x) \cdot \hat{z}}{f_x} + t_x, \quad Z = \hat{z}.
    \label{eq:unproject}
\end{equation}
With the proposal's predicted size, this gives a 4D BEV anchor $(X,Z,W,L)$ that initializes the BEV decoder, normalized within the grid extent ($X \in [-30, 30]$\,m, $Z \in [0, 60]$\,m). The seed depth $\hat{z}$ is the image branch's estimate (Section~\ref{sec:det2d}). Single-shot depth is unreliable (Section~\ref{sec:multiheight_ipm}), so the anchor is only \emph{initial}: the multi-height cross-attention re-estimates $Z$ per layer, allowing it to correct the seed using BEV evidence. Both decoders share one set of $N_q$ queries, with BEV query $i$ a learned projection of the final image query $i$ at the same proposal. Detaching the queries and anchor separates the decoder gradients, and the BEV decoder refines the box toward wherever the multi-height evidence supports it.

\subsubsection{BEV Transformer Decoder.}
The BEV decoder is a stack of $N_\mathrm{bev}$ layers, each applying self-attention among queries, multi-scale deformable cross-attention over the multi-height BEV features, and a feed-forward network. The $K$ height levels are treated as separate ``scales'' in deformable attention~\cite{zhu2021deformable}: each query samples $n_\mathrm{points}$ offsets from \emph{every} level and softmax-weights all $K{\times}n_\mathrm{points}$ samples into one output. Each level samples a \emph{different image row} for the same BEV cell (Eq.~\eqref{eq:vertical_shift}), so the fused evidence is height-conditioned with depth-dependent $1/Z$ spacing, and deformable attention learns which levels are most useful per query.

Following DAB-DETR~\cite{liu2022dabdetr}, each query's positional embedding is \emph{generated} from its 4D anchor $(X,Z,W,L)$ by sinusoidal encoding and a small MLP, and added to the content query in self- and cross-attention with the anchor center as the deformable reference; encoding $(W,L)$ makes the prior \emph{size-aware}. Anchors are initialized from the unprojected 2D proposals (Eq.~\eqref{eq:unproject}) rather than learned from scratch, a geometric warm start that stabilizes optimization. At each layer the BEV box head (a small MLP) regresses a residual update to the axis-aligned box $(X,Z,W,L)$ and a bounded residual to its log-scale $\log\sigma_z$, initialized from the depth head; heading is predicted from the BEV decoder features by a separate 12-bin/residual angle head, supervised at every BEV layer. Anchor and $\log\sigma_z$ are updated and detached between layers, so both are \emph{iteratively refined} under deep supervision and the regenerated positional query tracks the improving estimate. At inference, image-predicted centers are unprojected using the BEV-refined depth. The final dimensions retain image-predicted height and BEV-refined width/length. We select top-50 detections by class score and export uncertainty-weighted scores $s_\mathrm{cls}\cdot\exp(-\log\sigma_z){=}s_\mathrm{cls}/\sigma_z$, without NMS. Both decoders train jointly end-to-end.

\subsection{Loss Function}
\label{sec:loss}

Following Group DETR~\cite{chen2023groupdetr}, we replicate queries into $G$ groups during training, each independently matched via the Hungarian algorithm~\cite{kuhn1955hungarian}. This increases effective supervision without changing inference (single group of $N_q$ queries).

\subsubsection{BEV-Aware Matching Cost.}
Standard DETR matchers assign each ground-truth object to the query that is closest in the \emph{image plane} (classification score, 2D box $\ell_1$, and GIoU). Our BEV decoder, however, localizes objects in metric space, so an image-plane assignment can supervise a query whose perspective box overlaps the target but whose \emph{metric} BEV location is wrong, exactly the failure foreshortening induces. We therefore add a \emph{BEV-center} term to the matching cost: the $\ell_1$ distance between predicted and ground-truth object centers $(X,Z)$ in the normalized BEV grid. The total assignment cost is
\begin{equation}
    \mathcal{C} = \lambda_\mathrm{cls}\mathcal{C}_\mathrm{cls}
    + \lambda_\mathrm{box}\mathcal{C}_\mathrm{box}
    + \lambda_\mathrm{giou}\mathcal{C}_\mathrm{giou}
    + \lambda_\mathrm{3D}\mathcal{C}_\mathrm{3D}
    + \lambda_\mathrm{BEV}\mathcal{C}_\mathrm{BEV},
    \label{eq:matchcost}
\end{equation}
where $\mathcal{C}_\mathrm{3D}$ is the $\ell_1$ on the projected 3D center and $\mathcal{C}_\mathrm{cls},\mathcal{C}_\mathrm{box},\mathcal{C}_\mathrm{giou}$ are the standard DETR terms, with weights $(\lambda_\mathrm{cls},\lambda_\mathrm{box},\lambda_\mathrm{giou},\lambda_\mathrm{3D},\lambda_\mathrm{BEV}){=}(2,5,2,10,100)$. The large BEV coefficient compensates for the small normalized-grid distances and favors metric consistency alongside the image-plane terms.

The total loss combines both decoders with deep supervision at each layer:
\begin{equation}
    \mathcal{L}
    = \sum_{i=0}^{N_\mathrm{dec}-1} \mathcal{L}_\mathrm{3D}^{(i)}
    + \sum_{j=0}^{N_\mathrm{bev}-1} \mathcal{L}_\mathrm{BEV}^{(j)}
    + \mathcal{L}_\mathrm{dmap}
    + \mathcal{L}_\mathrm{bmap}.
    \label{eq:total_loss}
\end{equation}

The 3D detection loss $\mathcal{L}_\mathrm{3D}$ comprises sigmoid focal loss~\cite{lin2017focal} for classification, L1 + GIoU for 2D boxes, L1 for projected 3D center and dimensions, and Laplacian aleatoric uncertainty loss~\cite{kendall2017uncertainties} for depth:
\begin{equation}
    \mathcal{L}_\mathrm{depth} = \frac{\sqrt{2}}{\sigma_z} |z - \hat{z}| + \log(\sigma_z),
\end{equation}
which jointly learns the depth estimate $\hat{z}$ and its uncertainty $\sigma_z$.

The BEV loss $\mathcal{L}_\mathrm{BEV}$ includes classification, dimensions, depth, normalized BEV-center $\ell_1$, and angle-bin classification plus residual regression. The retained image-box fields also receive losses under the BEV-stage matching. The BEV footprint uses an axis-aligned $(X,Z,W,L)$ parameterization, with center and dimensions supervised separately from heading. Heading is predicted and supervised at every BEV layer, not by $\mathcal{L}_\mathrm{3D}$. GIoU is applied to the axis-aligned image boxes. $\mathcal{L}_\mathrm{dmap}$ and $\mathcal{L}_\mathrm{bmap}$ are the dense depth-map and BEV occupancy (BCE + Dice) losses.

\section{Experiments}
\label{sec:experiments}

\subsection{Experimental Setup}

\subsubsection{Dataset and Metrics.}
We evaluate on the KITTI 3D benchmark~\cite{geiger2012kitti} (7,481 training, 7,518 testing images; Car, Pedestrian, Cyclist), splitting training into 3,712 train / 3,769 val~\cite{chen20153d}; objects are graded Easy/Moderate/Hard by 2D box height, occlusion, and truncation. We additionally evaluate on the Waymo Open Dataset~\cite{sun2020scalability}, using the front camera and the DEVIANT~\cite{kumar2022deviant} split and protocol~\suppref{Sec.~H}. Input images are resized to $1280\times384$. Following the augmentation protocol of MonoDGP~\cite{pu2025monodgp}, we apply photometric distortion, random horizontal flip, and random crop/scale (each with probability $0.5$ and a $\pm5\%$ range), together with Mixup3D copy-paste~\cite{li2024monolss}. Part of our supervision lives in metric BEV space, so augmentation is calibration-consistent: flip and crop transform the camera matrix $\mathbf{P}_2$, and the BEV-center targets are re-derived from the augmented $\mathbf{P}_2$ so that image-space and BEV supervision stay geometrically aligned. We report $\mathrm{AP_{3D}}$ and $\mathrm{AP_{BEV}}$ at 40 recall points (R40)~\cite{simonelli2019disentangling}, ranking by Moderate $\mathrm{AP_{3D}}$ for Car at IoU$=$0.7.

\subsubsection{Implementation Details.}
We use ResNet-50~\cite{he2016resnet} for Car-only and all-class models. We train for $250$ epochs on a single NVIDIA RTX 4090 with AdamW~\cite{loshchilov2019adamw} (learning rate $2{\times}10^{-4}$, weight decay $10^{-4}$, batch size $8$), stepping the learning rate down at epochs $\{85,125,165,205\}$. At inference, class scores select top-50 predictions and exported scores use $s_\mathrm{cls}/\sigma_z$, without NMS. The default confidence threshold is $0.2$. Full details appear in the supplement~\suppref{Sec.~A, Tab.~S1}.

\begin{widetable}
\centering
\caption{\textbf{KITTI Car comparison at IoU$=$0.7.} R40 3D/BEV AP on \textit{val} and official \textit{test}. \textbf{Bold}/\underline{underline}: best/second-best without extra sensor data. $^\ddagger$: Car-only; $^\dagger$: all-class training (Pedestrian/Cyclist in Table~\ref{tab:multiclass_eff}). MonoIA also uses language/foundation-model conditioning.}
\label{tab:main_val}
\scriptsize
\setlength{\tabcolsep}{1.1pt}
\resizebox{\textwidth}{!}{\begin{tabular}{@{}lcc|rrr|rrr|rrr|rrr@{}}
\toprule
Method & \shortstack{Extra\\data} & Venue & \multicolumn{3}{c|}{Val. $\mathrm{AP_{3D|R40}}$} & \multicolumn{3}{c|}{Val. $\mathrm{AP_{BEV|R40}}$} & \multicolumn{3}{c|}{Test $\mathrm{AP_{3D|R40}}$} & \multicolumn{3}{c}{Test $\mathrm{AP_{BEV|R40}}$} \\
 & & & Easy & Mod. & Hard & Easy & Mod. & Hard & Easy & Mod. & Hard & Easy & Mod. & Hard \\
\midrule
CaDDN~\cite{reading2021caddn} & LiDAR & CVPR'21 & 23.57 & 16.31 & 13.84 & -- & -- & -- & 19.17 & 13.41 & 11.46 & 27.94 & 18.91 & 17.19 \\
\shortstack[l]{OccupancyM3D\\\cite{peng2024occupancym3d}} & & CVPR'24 & 26.87 & 19.96 & 17.15 & 35.72 & 26.60 & 23.68 & 25.55 & 17.02 & 14.79 & 35.38 & 24.18 & 21.37 \\
\cmidrule(lr){1-15}
OPA-3D~\cite{su2023opa3d} & Depth & ICRA'23 & 24.97 & 19.40 & 16.59 & 33.80 & 25.51 & 22.13 & 24.60 & 17.05 & 14.25 & 33.54 & 22.53 & 19.22 \\
\midrule
MonoFlex~\cite{zhang2021monoflex} & None & CVPR'21 & 23.64 & 17.51 & 14.83 & -- & -- & -- & 19.94 & 13.89 & 12.07 & 28.23 & 19.75 & 16.89 \\
GUPNet~\cite{lu2021gupnet} & & ICCV'21 & 22.76 & 16.46 & 13.72 & 31.07 & 22.94 & 19.75 & 20.11 & 14.20 & 11.77 & -- & -- & -- \\
MonoDDE~\cite{li2022monodde} & & CVPR'22 & 26.66 & 19.75 & 16.72 & 35.51 & 26.48 & 23.07 & 24.93 & 17.14 & 15.10 & 33.58 & 23.46 & 20.37 \\
MonoCon~\cite{liu2022monocon} & & AAAI'22 & 26.33 & 19.01 & 15.98 & -- & -- & -- & 22.50 & 16.46 & 13.95 & 31.12 & 22.10 & 19.00 \\
DEVIANT~\cite{kumar2022deviant} & & ECCV'22 & 24.63 & 16.54 & 14.52 & 32.60 & 23.04 & 19.99 & 21.88 & 14.46 & 11.89 & 29.65 & 20.44 & 17.43 \\
MonoDETR~\cite{zhang2023monodetr} & & ICCV'23 & 28.84 & 20.61 & 16.38 & 37.86 & 26.95 & 22.80 & 25.00 & 16.47 & 13.58 & 33.60 & 22.11 & 18.60 \\
MonoCD~\cite{yan2024monocd} & & CVPR'24 & 26.45 & 19.37 & 16.38 & 34.60 & 24.96 & 21.51 & 25.53 & 16.59 & 14.53 & 33.41 & 22.81 & 19.57 \\
FD3D~\cite{wu2024fd3d} & & AAAI'24 & 28.22 & 20.23 & 17.04 & 36.98 & 26.77 & 23.16 & 25.38 & 17.12 & 14.50 & 34.20 & 23.72 & 20.76 \\
MonoMAE~\cite{jiang2024monomae} & & NeurIPS'24 & 30.29 & 20.90 & 17.61 & 40.26 & 27.08 & 23.14 & 25.60 & 18.84 & 16.78 & 34.15 & 24.93 & 21.76 \\
MonoDGP~\cite{pu2025monodgp} & & CVPR'25 & 30.76 & 22.34 & 19.02 & 39.40 & 28.20 & 24.42 & 26.35 & 18.72 & 15.97 & 35.24 & 25.23 & 22.02 \\
MonoCoP~\cite{zhang2026monocop} & & CVPR'26 & 32.20 & 23.72 & 20.31 & 41.00 & 29.67 & 25.90 & 27.54 & 19.11 & 16.33 & 36.77 & 25.75 & 22.62 \\
MonoIA~\cite{zhang2026monoia} & & CVPR'26 & \textbf{33.61} & 24.40 & 20.80 & \textbf{44.69} & 32.17 & 27.93 & \underline{29.52} & \underline{20.29} & \underline{17.93} & \underline{37.55} & \underline{26.59} & \underline{23.26} \\
\midrule
\ours{}$^\ddagger$ & None & -- & \underline{33.24} & \textbf{25.84} & \textbf{23.17} & \underline{42.09} & \underline{32.66} & \textbf{29.90} & \textbf{30.85} & \textbf{21.66} & \textbf{18.77} & \textbf{37.80} & \textbf{27.85} & \textbf{24.81} \\
\ours{}$^\dagger$ & & -- & 32.85 & \underline{25.63} & \underline{22.46} & 41.51 & \textbf{32.78} & \underline{29.34} & 29.42 & 20.19 & 17.68 & 36.52 & 26.44 & 23.20 \\
\bottomrule
\end{tabular}}
\end{widetable}

\subsection{Main Results}

On the primary $\mathrm{AP_{3D}}$ Moderate metric (Table~\ref{tab:main_val}), \ours{} reaches $25.84\%$ on \textit{val} and $21.66\%$ on the official \textit{test} set with Car-only training, surpassing recent published monocular baselines in the table, including MonoDGP, MonoCoP, and MonoIA. The all-class model obtains $25.63\%$ $\mathrm{AP_{3D}}$ Moderate on \textit{val}, showing that the representation remains effective without category-specific tuning. \ours{} is especially strong on Moderate/Hard settings, indicating that BEV features improve metric localization under depth ambiguity and occlusion.

\subsubsection{Multi-Class Results and Efficiency.}
Trained jointly on all categories (Table~\ref{tab:multiclass_eff}), \ours{} achieves the best Pedestrian AP ($+2.08$ over MonoCoP~\cite{zhang2026monocop}, $+2.52$ over MonoDGP) and the best Cyclist AP ($+0.93$ over MonoCoP), with no category-specific design. Although the default planes span a car-scale height range (Section~\ref{sec:multiheight_ipm}), the representation is class-shared: every category uses the same reference and height-conditioned evidence. This does not imply that a finite plane range covers arbitrary object geometries.

\begin{widetable}
\centering
\caption{\textbf{Multi-class detection and efficiency.} Pedestrian and Cyclist KITTI \textit{test} $\mathrm{AP_{3D|R40}}$ (Moderate, IoU$=$0.5; all classes trained jointly, no extra data); we \textbf{bold} the best and \underline{underline} the second-best AP. Parameters and inference time are measured by us for all methods under one protocol on a single RTX~3090 (ResNet-50 backbone, batch size 1, forward only, FP32).}
\label{tab:multiclass_eff}
\setlength{\tabcolsep}{8pt}
\small
\begin{tabular}{l|cc|cc}
\toprule
Method & Ped.\ (Mod.) & Cyc.\ (Mod.) & Params (M) & Time (ms) \\
\midrule
MonoDGP~\cite{pu2025monodgp} & 9.89 & 2.82 & 42.16 & 25.6 \\
MonoCoP~\cite{zhang2026monocop} & \underline{10.33} & \underline{5.08} & 42.53 & 25.3 \\
\midrule
\ours{} (Ours) & \textbf{12.41} & \textbf{6.01} & 39.86 & 27.9 \\
\bottomrule
\end{tabular}
\end{widetable}

\begin{figure}[t]
\centering
\includegraphics[width=\textwidth]{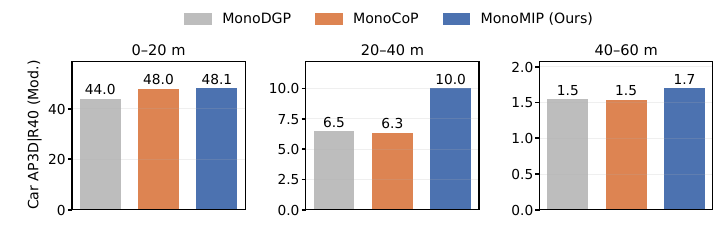}
\caption{\textbf{Per-distance Car $\mathrm{AP_{3D|R40}}$ (Moderate), KITTI \textit{val}.} Panels use separate $y$-axis scales. Gains concentrate in the $20$--$40$\,m band, where cross-height row separation stays visible but monocular depth is hard.}
\label{fig:perdist}
\end{figure}

\subsection{Ablation Study}

\begin{table}[!t]
\centering
\caption{\textbf{Ablations on KITTI \textit{val}} (Car, $\mathrm{AP_{R40}}$; ResNet-50). \textbf{(a)}~More height levels $K$ improve AP. \textbf{(b)}~Components added to the base image-space decoder (IoU$=$0.7).}
\label{tab:ablations}
\setlength{\tabcolsep}{3.75pt}
\footnotesize
\begin{tabular}{c|l|ccc|ccc}
\multicolumn{8}{c}{\textbf{(a) Number of height levels $K$}}\\[2pt]
\toprule
$K$ & Heights (m) & \multicolumn{3}{c|}{$\mathrm{AP_{3D|R40}}$} & \multicolumn{3}{c}{$\mathrm{AP_{BEV|R40}}$} \\
 & & Easy & Mod. & Hard & Easy & Mod. & Hard \\
\midrule
1 & \{1.65\} & 32.83 & 24.83 & 21.78 & 41.26 & 31.18 & 27.65 \\ 2 & \{0.25, 1.65\} & 33.04 & 25.34 & 22.46 & 41.68 & 31.92 & 28.78 \\
3 & \{0.25, 1.0, 1.65\} & 33.16 & 25.64 & 22.89 & 41.92 & 32.36 & 29.45 \\
4 & \{0.25, 0.75, 1.25, 1.65\} & \textbf{33.24} & \textbf{25.84} & \textbf{23.17} & \textbf{42.09} & \textbf{32.66} & \textbf{29.90} \\
\bottomrule
\end{tabular}

\begin{tabular}{ccc|ccc|ccc}
\multicolumn{9}{c}{\textbf{(b) Component ablation}}\\[2pt]
\toprule
MH-IPM & BEV Dec. & Gauss. Depth & \multicolumn{3}{c|}{$\mathrm{AP_{3D|R40}}$} & \multicolumn{3}{c}{$\mathrm{AP_{BEV|R40}}$} \\
 & & & Easy & Mod. & Hard & Easy & Mod. & Hard \\
\midrule
$\times$ & $\times$ & $\times$ & 29.58 & 21.16 & 17.20 & 38.49 & 27.66 & 23.38 \\
$\times$ & $\times$ & $\checkmark$ & 31.33 & 23.01 & 19.23 & 40.24 & 29.60 & 25.45 \\
$\checkmark$ & $\checkmark$ & $\times$ & 32.76 & 25.27 & 22.65 & 41.58 & 32.08 & 29.29 \\
$\checkmark$ & $\checkmark$ & $\checkmark$ & \textbf{33.24} & \textbf{25.84} & \textbf{23.17} & \textbf{42.09} & \textbf{32.66} & \textbf{29.90} \\
\bottomrule
\end{tabular}

\end{table}

Table~\ref{tab:ablations} separates the contributions. \textbf{(a)}~Adding height planes improves AP with diminishing returns: $K{=}1\to4$ gives $24.83\to25.84$, a $1.01$ AP gain. \textbf{(b)}~The full $4.68$ AP gain over the $21.16$ image-space baseline combines the BEV pathway and depth targets, not multiple planes alone. Gaussian targets improve the image baseline ($21.16\to23.01$) and add a smaller complementary gain with BEV ($25.27\to25.84$). Additional four-plane coverage variants reach $25.30/24.33$ AP without nominal ground ($1.65$\,m), supporting suitable vertical coverage rather than mandatory ground coincidence~\suppref{Sec.~E}. Per-distance gains concentrate at $20$--$40$\,m (Figure~\ref{fig:perdist}), consistent with the $1/Z$ geometry.

\paragraph{Controlled lifting comparison.}
An LSS-style replacement keeps the fused features, depth predictor, BEV grid, queries, decoder, losses, data, and schedule fixed. Its three-seed mean and maximum are $24.05$ and $24.17$, below all ten MH-IPM runs ($24.80$--$25.84$). MH-IPM takes $28.14$\,ms versus $30.55$\,ms for LSS, a $7.9\%$ reduction. Decoder-composition results and measurement details are in the supplement~\suppref{Sec.~K}.

\section{Conclusion}
\label{sec:conclusion}

We presented \ours{}, a monocular 3D detector pairing depth-guided image-space and metric BEV reasoning from multi-height IPM. Projecting features to fixed planes reduces above-ground misalignment and exposes ordered, depth-dependent evidence the BEV decoder aggregates into metric centers, improving monocular 3D localization on both KITTI and Waymo.

\subsubsection{Limitations and Future Work.}
\ours{} assumes known calibration and fixed height planes, with far-range evidence limited by feature stride~\suppref{Sec.~J}. Matched focal tests attribute most sensitivity to geometric depth, while camera-height error remains an MH-IPM limitation. Pitch and non-planar roads remain unvalidated. A $192{\times}160$ grid improves Cyclist by $0.79$ AP but Pedestrian by only $0.12$, so resolution is one factor rather than a complete explanation of small-object errors. Reported Waymo results retain KITTI-default planes~\suppref{Sec.~H}. Scene-adaptive heights and uncertainty-aware calibration are future directions.

\FloatBarrier
\label{paper-body-end}
\clearpage
\bibliographystyle{splncs04}
\bibliography{main_references}

\begin{thebibliography}{10}
\providecommand{\url}[1]{\texttt{#1}}
\providecommand{\urlprefix}{URL }
\providecommand{\doi}[1]{https://doi.org/#1}

\bibitem{brazil2019m3drpn}
Brazil, G., Liu, X.: {M3D-RPN: Monocular 3D Region Proposal Network for Object
  Detection}. In: IEEE/CVF International Conference on Computer Vision (ICCV).
  pp. 9287--9296 (2019)

\bibitem{carion2020detr}
Carion, N., Massa, F., Synnaeve, G., Usunier, N., Kirillov, A., Zagoruyko, S.:
  {End-to-End Object Detection with Transformers}. In: European Conference on
  Computer Vision (ECCV). pp. 213--229 (2020)

\bibitem{chen2023groupdetr}
Chen, Q., Chen, X., Wang, J., Zhang, S., Yao, K., Feng, H., Han, J., Ding, E.,
  Zeng, G., Wang, J.: {Group DETR: Fast DETR Training with Group-Wise
  One-to-Many Assignment}. In: IEEE/CVF International Conference on Computer
  Vision (ICCV). pp. 6633--6642 (2023)

\bibitem{chen20153d}
Chen, X., Kundu, K., Zhu, Y., Berneshawi, A.G., Ma, H., Fidler, S., Urtasun,
  R.: {3D Object Proposals for Accurate Object Class Detection}. In: Advances
  in Neural Information Processing Systems (NeurIPS). vol.~28 (2015)

\bibitem{chen2020dsgn}
Chen, Y., Liu, S., Shen, X., Jia, J.: {DSGN: Deep Stereo Geometry Network for
  3D Object Detection}. In: IEEE/CVF Conference on Computer Vision and Pattern
  Recognition (CVPR) (2020)

\bibitem{garnett2019lanenet}
Garnett, N., Cohen, R., Pe'er, T., Lahav, R., Levi, D.: {3D-LaneNet: End-to-End
  3D Multiple Lane Detection}. In: IEEE/CVF International Conference on
  Computer Vision (ICCV) (2019)

\bibitem{geiger2012kitti}
Geiger, A., Lenz, P., Urtasun, R.: {Are We Ready for Autonomous Driving? The
  KITTI Vision Benchmark Suite}. In: IEEE Conference on Computer Vision and
  Pattern Recognition (CVPR). pp. 3354--3361 (2012)

\bibitem{guo2021ligastereo}
Guo, X., Shi, S., Wang, X., Li, H.: {LIGA-Stereo: Learning LiDAR Geometry Aware
  Representations for Stereo-Based 3D Detector}. In: IEEE/CVF International
  Conference on Computer Vision (ICCV). pp. 3153--3163 (2021)

\bibitem{he2016resnet}
He, K., Zhang, X., Ren, S., Sun, J.: {Deep Residual Learning for Image
  Recognition}. In: IEEE Conference on Computer Vision and Pattern Recognition
  (CVPR). pp. 770--778 (2016)

\bibitem{huang2021bevdet}
Huang, J., Huang, G., Zhu, Z., Ye, Y., Du, D.: {BEVDet: High-Performance
  Multi-Camera 3D Object Detection in Bird-Eye-View}. arXiv preprint
  arXiv:2112.11790  (2021)

\bibitem{huang2022monodtr}
Huang, K.C., Wu, T.H., Su, H.T., Hsu, W.H.: {MonoDTR: Monocular 3D Object
  Detection with Depth-Aware Transformer}. In: IEEE Conference on Computer
  Vision and Pattern Recognition (CVPR). pp. 4012--4021 (2022)

\bibitem{jiang2024monomae}
Jiang, X., Jin, S., Zhang, X., Shao, L., Lu, S.: {MonoMAE: Enhancing Monocular
  3D Detection through Depth-Aware Masked Autoencoders}. In: Advances in Neural
  Information Processing Systems (NeurIPS) (2024)

\bibitem{kendall2017uncertainties}
Kendall, A., Gal, Y.: {What Uncertainties Do We Need in Bayesian Deep Learning
  for Computer Vision?} In: Advances in Neural Information Processing Systems
  (NeurIPS). vol.~30 (2017)

\bibitem{kuhn1955hungarian}
Kuhn, H.W.: {The Hungarian Method for the Assignment Problem}. Naval Research
  Logistics Quarterly  \textbf{2}(1-2),  83--97 (1955)

\bibitem{kumar2022deviant}
Kumar, A., Brazil, G., Corona, E., Parchami, A., Liu, X.: {DEVIANT: Depth
  Equivariant Network for Monocular 3D Object Detection}. In: European
  Conference on Computer Vision (ECCV). pp. 664--683 (2022)

\bibitem{lang2019pointpillars}
Lang, A.H., Vora, S., Caesar, H., Zhou, L., Yang, J., Beijbom, O.:
  {PointPillars: Fast Encoders for Object Detection from Point Clouds}. In:
  IEEE Conference on Computer Vision and Pattern Recognition (CVPR). pp.
  12697--12705 (2019)

\bibitem{li2022bevstereo}
Li, Y., Bao, H., Ge, Z., Yang, J., Sun, J., Li, Z.: {BEVStereo: Enhancing Depth
  Estimation in Multi-view 3D Object Detection with Temporal Stereo}. In: AAAI
  Conference on Artificial Intelligence. pp. 1486--1494 (2023)

\bibitem{li2023bevdepth}
Li, Y., Ge, Z., Yu, G., Yang, J., Wang, Z., Shi, Y., Sun, J., Li, Z.:
  {BEVDepth: Acquisition of Reliable Depth for Multi-view 3D Object Detection}.
  In: AAAI Conference on Artificial Intelligence. pp. 1477--1485 (2023)

\bibitem{li2024monolss}
Li, Z., Jia, J., Shi, Y.: {MonoLSS: Learnable Sample Selection for Monocular 3D
  Detection}. In: International Conference on 3D Vision (3DV). pp. 1125--1135
  (2024)

\bibitem{li2022bevformer}
Li, Z., Wang, W., Li, H., Xie, E., Sima, C., Lu, T., Qiao, Y., Dai, J.:
  {BEVFormer: Learning Bird's-Eye-View Representation from Multi-Camera Images
  via Spatiotemporal Transformers}. In: European Conference on Computer Vision
  (ECCV). pp. 1--18 (2022)

\bibitem{li2022monodde}
Li, Z., Qu, Z., Zhou, Y., Liu, J., Wang, H., Jiang, L.: {Diversity Matters:
  Fully Exploiting Depth Clues for Reliable Monocular 3D Object Detection}. In:
  IEEE Conference on Computer Vision and Pattern Recognition (CVPR). pp.
  2791--2800 (2022)

\bibitem{lin2017focal}
Lin, T.Y., Goyal, P., Girshick, R., He, K., Doll{\'a}r, P.: {Focal Loss for
  Dense Object Detection}. In: IEEE/CVF International Conference on Computer
  Vision (ICCV). pp. 2980--2988 (2017)

\bibitem{liu2022dabdetr}
Liu, S., Li, F., Zhang, H., Yang, X., Qi, X., Su, H., Zhu, J., Zhang, L.:
  {DAB-DETR: Dynamic Anchor Boxes are Better Queries for DETR}. In:
  International Conference on Learning Representations (ICLR) (2022)

\bibitem{liu2022monocon}
Liu, X., Xue, N., Wu, T.: {Learning Auxiliary Monocular Contexts Helps
  Monocular 3D Object Detection}. In: AAAI Conference on Artificial
  Intelligence. pp. 1810--1818 (2022)

\bibitem{liu2023monoxiver}
Liu, X., Zheng, C., Cheng, K., Xue, N., Qi, G.J., Wu, T.: {Monocular 3D Object
  Detection with Bounding Box Denoising in 3D by Perceiver}. In: IEEE/CVF
  International Conference on Computer Vision (ICCV) (2023)

\bibitem{liu2022petr}
Liu, Y., Wang, T., Zhang, X., Sun, J.: {PETR: Position Embedding Transformation
  for Multi-View 3D Object Detection}. In: European Conference on Computer
  Vision (ECCV). pp. 531--548 (2022)

\bibitem{liu2020smoke}
Liu, Z., Wu, Z., T{\'o}th, R.: {SMOKE: Single-Stage Monocular 3D Object
  Detection via Keypoint Estimation}. In: IEEE/CVF Conference on Computer
  Vision and Pattern Recognition Workshops (CVPRW). pp. 996--997 (2020)

\bibitem{loshchilov2019adamw}
Loshchilov, I., Hutter, F.: {Decoupled Weight Decay Regularization}. In:
  International Conference on Learning Representations (ICLR) (2019)

\bibitem{lu2021gupnet}
Lu, Y., Ma, X., Yang, L., Zhang, T., Liu, Y., Chu, Q., Yan, J., Ouyang, W.:
  {Geometry Uncertainty Projection Network for Monocular 3D Object Detection}.
  In: IEEE/CVF International Conference on Computer Vision (ICCV). pp.
  3111--3121 (2021)

\bibitem{ma2021monodle}
Ma, X., Zhang, Y., Xu, D., Zhou, D., Yi, S., Li, H., Ouyang, W.: {Delving into
  Localization Errors for Monocular 3D Object Detection}. In: IEEE Conference
  on Computer Vision and Pattern Recognition (CVPR). pp. 4721--4730 (2021)

\bibitem{mallot1991ipm}
Mallot, H.A., B{\"u}lthoff, H.H., Little, J.J., Bohrer, S.: {Inverse
  Perspective Mapping Simplifies Optical Flow Computation and Obstacle
  Detection}. Biological Cybernetics  \textbf{64}(3),  177--185 (1991)

\bibitem{park2021pseudolidar}
Park, D., Ambrus, R., Guizilini, V., Li, J., Gaidon, A.: {Is Pseudo-Lidar
  Needed for Monocular 3D Object Detection?} In: IEEE/CVF International
  Conference on Computer Vision (ICCV). pp. 3142--3152 (2021)

\bibitem{peng2024occupancym3d}
Peng, L., Xu, J., Cheng, H., Yang, Z., Wu, X., Qian, W., Wang, W., Wu, B., Cai,
  D.: {Learning Occupancy for Monocular 3D Object Detection}. In: IEEE
  Conference on Computer Vision and Pattern Recognition (CVPR). pp.
  10281--10292 (2024)

\bibitem{philion2020lift}
Philion, J., Fidler, S.: {Lift, Splat, Shoot: Encoding Images from Arbitrary
  Camera Rigs by Implicitly Unprojecting to 3D}. In: European Conference on
  Computer Vision (ECCV). pp. 194--210 (2020)

\bibitem{pu2025monodgp}
Pu, F., Wang, Y., Deng, J., Yang, W.: {MonoDGP: Monocular 3D Object Detection
  with Decoupled-Query and Geometry-Error Priors}. In: IEEE Conference on
  Computer Vision and Pattern Recognition (CVPR) (2025)

\bibitem{ranasinghe2024monodiff}
Ranasinghe, Y., Hegde, D., Patel, V.M.: {MonoDiff: Monocular 3D Object
  Detection and Pose Estimation with Diffusion Models}. In: IEEE Conference on
  Computer Vision and Pattern Recognition (CVPR). pp. 10659--10670 (2024)

\bibitem{reading2021caddn}
Reading, C., Harakeh, A., Chae, J., Waslander, S.L.: {Categorical Depth
  Distribution Network for Monocular 3D Object Detection}. In: IEEE Conference
  on Computer Vision and Pattern Recognition (CVPR). pp. 8555--8564 (2021)

\bibitem{reiher2020sim2real}
Reiher, L., Lampe, B., Eckstein, L.: {A Sim2Real Deep Learning Approach for the
  Transformation of Images from Multiple Vehicle-Mounted Cameras to a
  Semantically Segmented Image in Bird's Eye View}. In: IEEE International
  Conference on Intelligent Transportation Systems (ITSC). pp.~1--7 (2020)

\bibitem{roddick2019oft}
Roddick, T., Kendall, A., Cipolla, R.: {Orthographic Feature Transform for
  Monocular 3D Object Detection}. In: British Machine Vision Conference (BMVC)
  (2019)

\bibitem{shi2019pointrcnn}
Shi, S., Wang, X., Li, H.: {PointRCNN: 3D Object Proposal Generation and
  Detection from Point Cloud}. In: IEEE Conference on Computer Vision and
  Pattern Recognition (CVPR). pp. 770--779 (2019)

\bibitem{shi2021geometry}
Shi, X., Ye, Q., Chen, X., Chen, C., Chen, Z., Kim, T.K.: {Geometry-based
  Distance Decomposition for Monocular 3D Object Detection}. In: IEEE/CVF
  International Conference on Computer Vision (ICCV). pp. 15172--15181 (2021)

\bibitem{simonelli2019disentangling}
Simonelli, A., Rota~Bul{\`o}, S., Porzi, L., L{\'o}pez-Antequera, M.,
  Kontschieder, P.: {Disentangling Monocular 3D Object Detection}. In: IEEE/CVF
  International Conference on Computer Vision (ICCV). pp. 1991--1999 (2019)

\bibitem{song2021shot}
Song, L., Wu, J., Yang, M., Zhang, Q., Li, Y., Yuan, J.: {Stacked Homography
  Transformations for Multi-View Pedestrian Detection}. In: Proceedings of the
  IEEE/CVF International Conference on Computer Vision (ICCV). pp. 6049--6057
  (2021)

\bibitem{su2023opa3d}
Su, Y., Di, Y., Zhai, G., Manhardt, F., Rambach, J., Busam, B., Stricker, D.,
  Tombari, F.: {OPA-3D: Occlusion-Aware Pixel-Wise Aggregation for Monocular 3D
  Object Detection}. IEEE Robotics and Automation Letters  \textbf{8}(3),
  1327--1334 (2023)

\bibitem{sun2020scalability}
Sun, P., Kretzschmar, H., Dotiwalla, X., Chouard, A., Patnaik, V., Tsui, P.,
  Guo, J., Zhou, Y., Chai, Y., Caine, B., Vasudevan, V., Han, W., Ngiam, J.,
  Zhao, H., Timofeev, A., Ettinger, S., Krivokon, M., Gao, A., Joshi, A.,
  Zhang, Y., Shlens, J., Chen, Z., Anguelov, D.: {Scalability in Perception for
  Autonomous Driving: Waymo Open Dataset}. In: Proceedings of the IEEE/CVF
  Conference on Computer Vision and Pattern Recognition (CVPR). pp. 2446--2454
  (2020)

\bibitem{wang2021fcos3d}
Wang, T., Zhu, X., Pang, J., Lin, D.: {FCOS3D: Fully Convolutional One-Stage
  Monocular 3D Object Detection}. In: IEEE/CVF International Conference on
  Computer Vision (ICCV) Workshops. pp. 913--922 (2021)

\bibitem{wang2022pgd}
Wang, T., Zhu, X., Pang, J., Lin, D.: {Probabilistic and Geometric Depth:
  Detecting Objects in Perspective}. In: Conference on Robot Learning (CoRL).
  pp. 1475--1485 (2022)

\bibitem{wang2019pseudolidar}
Wang, Y., Chao, W.L., Garg, D., Hariharan, B., Campbell, M., Weinberger, K.Q.:
  {Pseudo-LiDAR from Visual Depth Estimation: Bridging the Gap in 3D Object
  Detection for Autonomous Driving}. In: IEEE Conference on Computer Vision and
  Pattern Recognition (CVPR). pp. 8445--8453 (2019)

\bibitem{wu2024fd3d}
Wu, Z., Gan, Y., Wu, Y., Wang, R., Wang, X., Pu, J.: {FD3D: Exploiting
  Foreground Depth Map for Feature-Supervised Monocular 3D Object Detection}.
  In: AAAI Conference on Artificial Intelligence. pp. 6189--6197 (2024)

\bibitem{yan2024monocd}
Yan, L., Yan, P., Xiong, S., Xiang, X., Tan, Y.: {MonoCD: Monocular 3D Object
  Detection with Complementary Depths}. In: IEEE Conference on Computer Vision
  and Pattern Recognition (CVPR). pp. 10248--10257 (2024)

\bibitem{yang2021projecting}
Yang, W., Li, Q., Liu, W., Yu, Y., Ma, Y., He, S., Pan, J.: {Projecting Your
  View Attentively: Monocular Road Scene Layout Estimation via Cross-View
  Transformation}. In: IEEE Conference on Computer Vision and Pattern
  Recognition (CVPR). pp. 15536--15545 (2021)

\bibitem{zhang2023monodetr}
Zhang, R., Qiu, H., Wang, T., Guo, Z., Cui, Z., Qiao, Y., Li, H., Gao, P.:
  {MonoDETR: Depth-guided Transformer for Monocular 3D Object Detection}. In:
  IEEE/CVF International Conference on Computer Vision (ICCV). pp. 9155--9166
  (2023)

\bibitem{zhang2021monoflex}
Zhang, Y., Lu, J., Zhou, J.: {Objects Are Different: Flexible Monocular 3D
  Object Detection}. In: IEEE Conference on Computer Vision and Pattern
  Recognition (CVPR). pp. 3289--3298 (2021)

\bibitem{zhang2026monocop}
Zhang, Z., Kumar, A., Ganesan, G.C., Liu, X.: {Unleashing the Power of
  Chain-of-Prediction for Monocular 3D Object Detection}. In: IEEE/CVF
  Conference on Computer Vision and Pattern Recognition (CVPR) (2026)

\bibitem{zhang2026monoia}
Zhang, Z., Kumar, A., Liu, X.: {Towards Intrinsic-Aware Monocular 3D Object
  Detection}. In: IEEE/CVF Conference on Computer Vision and Pattern
  Recognition (CVPR) (2026)

\bibitem{zhou2019objects}
Zhou, X., Wang, D., Kr{\"a}henb{\"u}hl, P.: {Objects as Points}. arXiv preprint
  arXiv:1904.07850  (2019)

\bibitem{zhu2021deformable}
Zhu, X., Su, W., Lu, L., Li, B., Wang, X., Dai, J.: {Deformable DETR:
  Deformable Transformers for End-to-End Object Detection}. In: International
  Conference on Learning Representations (ICLR) (2021)

\end{thebibliography}


\begin{thebibliography}{10}
\providecommand{\url}[1]{\texttt{#1}}
\providecommand{\urlprefix}{URL }
\providecommand{\doi}[1]{https://doi.org/#1}

\bibitem{jinrang2023monouni}
Jia, J., Li, Z., Shi, Y.: {MonoUNI: A Unified Vehicle and Infrastructure-side
  Monocular 3D Object Detection Network with Sufficient Depth Clues}. In:
  Advances in Neural Information Processing Systems (NeurIPS) (2023)

\bibitem{kumar2022deviant}
Kumar, A., Brazil, G., Corona, E., Parchami, A., Liu, X.: {DEVIANT: Depth
  Equivariant Network for Monocular 3D Object Detection}. In: European
  Conference on Computer Vision (ECCV). pp. 664--683 (2022)

\bibitem{lu2021gupnet}
Lu, Y., Ma, X., Yang, L., Zhang, T., Liu, Y., Chu, Q., Yan, J., Ouyang, W.:
  {Geometry Uncertainty Projection Network for Monocular 3D Object Detection}.
  In: IEEE/CVF International Conference on Computer Vision (ICCV). pp.
  3111--3121 (2021)

\bibitem{philion2020lift}
Philion, J., Fidler, S.: {Lift, Splat, Shoot: Encoding Images from Arbitrary
  Camera Rigs by Implicitly Unprojecting to 3D}. In: European Conference on
  Computer Vision (ECCV). pp. 194--210 (2020)

\bibitem{pu2025monodgp}
Pu, F., Wang, Y., Deng, J., Yang, W.: {MonoDGP: Monocular 3D Object Detection
  with Decoupled-Query and Geometry-Error Priors}. In: IEEE Conference on
  Computer Vision and Pattern Recognition (CVPR) (2025)

\bibitem{reading2021caddn}
Reading, C., Harakeh, A., Chae, J., Waslander, S.L.: {Categorical Depth
  Distribution Network for Monocular 3D Object Detection}. In: IEEE Conference
  on Computer Vision and Pattern Recognition (CVPR). pp. 8555--8564 (2021)

\bibitem{sun2020scalability}
Sun, P., Kretzschmar, H., Dotiwalla, X., Chouard, A., Patnaik, V., Tsui, P.,
  Guo, J., Zhou, Y., Chai, Y., Caine, B., Vasudevan, V., Han, W., Ngiam, J.,
  Zhao, H., Timofeev, A., Ettinger, S., Krivokon, M., Gao, A., Joshi, A.,
  Zhang, Y., Shlens, J., Chen, Z., Anguelov, D.: {Scalability in Perception for
  Autonomous Driving: Waymo Open Dataset}. In: Proceedings of the IEEE/CVF
  Conference on Computer Vision and Pattern Recognition (CVPR). pp. 2446--2454
  (2020)

\bibitem{zhang2023monodetr}
Zhang, R., Qiu, H., Wang, T., Guo, Z., Cui, Z., Qiao, Y., Li, H., Gao, P.:
  {MonoDETR: Depth-guided Transformer for Monocular 3D Object Detection}. In:
  IEEE/CVF International Conference on Computer Vision (ICCV). pp. 9155--9166
  (2023)

\bibitem{zhang2026monocop}
Zhang, Z., Kumar, A., Ganesan, G.C., Liu, X.: {Unleashing the Power of
  Chain-of-Prediction for Monocular 3D Object Detection}. In: IEEE/CVF
  Conference on Computer Vision and Pattern Recognition (CVPR) (2026)

\bibitem{zhu2021deformable}
Zhu, X., Su, W., Lu, L., Li, B., Wang, X., Dai, J.: {Deformable DETR:
  Deformable Transformers for End-to-End Object Detection}. In: International
  Conference on Learning Representations (ICLR) (2021)

\end{thebibliography}

\end{document}


\title{\texorpdfstring{Supplementary Material for \\ \ours{}: Multi-Height Inverse Perspective Mapping \\ for Monocular 3D Object Detection}{Supplementary Material for MonoMIP: Multi-Height Inverse Perspective Mapping for Monocular 3D Object Detection}}
\titlerunning{Supplementary Material for \ours{}}

\author{Emir Ulurak\inst{1} \and Furkan K{\i}ra\c{c}\inst{1}}
\authorrunning{E. Ulurak and F. K{\i}ra\c{c}}
\institute{Ozyegin University, Istanbul, Turkey\\
\email{emir.ulurak@ozu.edu.tr, furkan.kirac@ozyegin.edu.tr}}

\maketitle

\section{Full Implementation Details}
\label{sec:impl}
Table~\ref{tab:impl} gives the full configuration of the \ours{} model.

\begin{table}[htbp]
\centering
\caption{\textbf{Full implementation details} of the \ours{} model.}
\label{tab:impl}
\footnotesize
\begin{tabular}{ll}
\toprule
Component & Setting \\
\midrule
Backbone & ResNet-50 \\
Input resolution & $1280\times384$ (W$\times$H) \\
Feature levels & $L=4$ \\
Hidden dimension & $256$ \\
Feed-forward width & $256$ \\
Attention heads & $8$ \\
Image encoder layers & $N_\mathrm{enc}=3$ \\
Image decoder layers & $N_\mathrm{dec}=3$ \\
BEV decoder layers & $N_\mathrm{bev}=3$ \\
Object queries & $N_q=50$ \\
Training groups & $G=11$ \\
BEV grid & $96\times80$ \\
BEV range & $[0,60]\times[-30,30]$ m \\
Height planes & $\{0.25,0.75,1.25,1.65\}$ m \\
Depth bins & $N_d=80$, LID \\
Optimizer & AdamW \\
Learning rate & $2\times10^{-4}$ \\
Weight decay & $10^{-4}$ \\
Batch size & $8$ \\
Epochs & $250$ \\
LR drops & $\{85,125,165,205\}$ \\
Augmentations & photometric, flip, crop/scale, Mixup3D, $\mathbf{P}_2$ calibration \\
Inference & top-50 predictions, no NMS, confidence threshold $0.2$ \\
\bottomrule
\end{tabular}
\end{table}

\noindent\textbf{Gaussian depth-target implementation.}
The targets in main-paper Section~3.4 are constructed on the $80\times24$
depth-logit grid, at stride 16 for the $1280\times384$ input. All spatial
coordinates and widths below are measured in depth-map cells. Box corners
are rounded outward and clipped to the grid before computing box width and
height. We use $\sigma_s=0.25$, $\sigma_{\min}=3$, and a $3\times3$ exact-depth
patch centered at the rounded projected 3D center and clipped to the box.
Overlapping targets are combined by a per-cell minimum depth.

For the categorical target, let $d$ denote the resulting metric target,
$d_{\min}=10^{-3}$\,m, $d_{\max}=60$\,m, and $N_d=80$. Its continuous LID
index and the normalized soft labels are
\[
\begin{gathered}
\delta=\frac{2(d_{\max}-d_{\min})}{N_d(N_d+1)},\qquad
\iota(d)=\frac{\sqrt{1+8(d-d_{\min})/\delta}-1}{2},\\[3pt]
q_j=\frac{\exp[-(j-\iota(d))^2/(2\sigma_b^2)]}
{\sum_{k=0}^{N_d-1}\exp[-(k-\iota(d))^2/(2\sigma_b^2)]},
\end{gathered}
\]
where $j=0,\ldots,N_d-1$ and $\sigma_b=1$ bin. Background targets
($d\ge d_{\max}-10^{-3}$\,m), out-of-range depths, and non-finite values
instead receive unit mass in the extra bin $N_d$. The implementation uses
soft-label cross-entropy, which equals the stated KL objective up to the
fixed target entropy and gives the same prediction gradients. Foreground
and background cells receive weights 13 and 1, respectively, with the
weighted loss averaged over all cells.

\noindent\textbf{Calibration-consistent augmentation.} We update
$\mathbf{P}_2$ for every geometric image augmentation so that it agrees with
the multi-height IPM projection (Section~\ref{sec:geom}). A
crop-and-scale acts on homogeneous pixels $\tilde{\mathbf{u}}=(u,v,1)^{\!\top}$
as an affine with scale $s$ and crop offset $(o_x,o_y)$,
\begin{equation}
\tilde{\mathbf{u}}'=\mathbf{A}\,\tilde{\mathbf{u}},\qquad
\mathbf{A}=\begin{bmatrix} s & 0 & -o_x\\ 0 & s & -o_y\\ 0 & 0 & 1\end{bmatrix},
\label{eq:supp_affine}
\end{equation}
and a horizontal flip mirrors columns, $u\mapsto W\!-\!1\!-\!u$ (re-solved as a
least-squares $\mathbf{P}_2'$ by SVD). As $\mathbf{P}_2$ maps a world point to a
pixel, the augmented calibration is the left composition
\begin{equation}
\mathbf{P}_2'=\mathbf{A}\,\mathbf{P}_2 .
\label{eq:supp_p2aug}
\end{equation}
The BEV-center targets are then re-derived through $\mathbf{P}_2'$. Un-projecting
column $u$ at depth $Z$ (Eq.~\eqref{eq:supp_proj}) gives $X=(u-c_x)\,Z/f_x+t_x$;
substituting $f_x'=s f_x$, $c_x'=s c_x-o_x$, and $u'=s u-o_x$ (with $t_x'=t_x$
under the rectified approximation $p_z=0$) yields
\begin{equation}
X'=\frac{u'-c_x'}{f_x'}\,Z+t_x'
 =\frac{s\,(u-c_x)}{s\,f_x}\,Z+t_x
 =X,
\label{eq:supp_invariance}
\end{equation}
so crop-and-scale leaves the metric BEV center \emph{invariant}, since the affine on
$\mathbf{P}_2$ exactly cancels the one on the pixel, while a flip maps
$X\!\mapsto\!-X$ and heading $\theta\!\mapsto\!\pi-\theta$. Because the IPM
sampling grid and the BEV-center targets are computed from the same
$\mathbf{P}_2'$, the height-plane features and their supervision remain
geometrically aligned under augmentation; photometric distortion and Mixup3D
leave $\mathbf{P}_2$ unchanged.

Unless noted, all ablation variants share this recipe; published baseline rows
are cited, not re-trained under it.

\section{Multi-Height IPM Geometry}
\label{sec:geom}
KITTI provides the rectified camera-2 projection matrix $P_2$:
\begin{equation}
P_2=
\begin{bmatrix}
f_x & 0 & c_x & p_x\\
0 & f_y & c_y & p_y\\
0 & 0 & 1 & p_z
\end{bmatrix}.
\end{equation}
We extract the rectified intrinsic parameters $f_x,f_y,c_x,c_y$ (the entries of
the $3\times3$ intrinsic matrix $\mathbf{K}$) and the small rectified camera
offsets
\begin{equation}
t_x=-\frac{p_x}{f_x}, \qquad
t_y=-\frac{p_y}{f_y}.
\end{equation}
The implementation neglects the small $p_z$ offset in the rectified projection.
For a BEV grid cell $(X_i,Z_j)$ on the $96{\times}80$ metric grid
(Table~\ref{tab:impl}) and height plane $Y_k$, the sampled image coordinate is
\begin{equation}
u_{ijk}=f_x\frac{X_i-t_x}{Z_j}+c_x,
\qquad
v_{ijk}=f_y\frac{Y_k-t_y}{Z_j}+c_y.
\label{eq:supp_proj}
\end{equation}
The height-plane BEV feature is obtained by bilinear sampling of the fused image
feature map $\mathbf{F}_\mathrm{fused}$:
\begin{equation}
\mathbf{B}_k(i,j)=
\mathrm{bilinear}\!\left(\mathbf{F}_\mathrm{fused},\,u_{ijk},\,v_{ijk}\right).
\label{eq:supp_bilinear}
\end{equation}
A BEV cell need not project inside the image: cells whose back-projection at
plane $Y_k$ lies behind the camera or outside the field of view produce sampling
coordinates beyond the image bounds. We therefore keep a per-level validity mask
\begin{equation}
M_k(i,j)=\mathbf{1}\!\left[\,0\le u_{ijk}<W \,\wedge\, 0\le v_{ijk}<H\,\right],
\label{eq:supp_mask}
\end{equation}
and zero out invalid cells, $\mathbf{B}_k(i,j)\leftarrow M_k(i,j)\,\mathbf{B}_k(i,j)$,
so out-of-view cells contribute no spurious evidence to the BEV decoder. This is the
``masked to zero'' rule referenced in the main paper; $W$ and $H$ are the
image width and height.
For a fixed BEV cell, the cross-height row offset between planes $Y_k$ and $Y_l$
is
\begin{equation}
\Delta v_{kl}(Z_j)
= v_{ijk}-v_{ijl}
= f_y\frac{Y_k-Y_l}{Z_j}.
\label{eq:supp_dv}
\end{equation}
The offset $t_y$ cancels in Eq.~\eqref{eq:supp_dv}, so the cross-height row
spacing preserves the $1/Z$ form used in the main paper. At feature stride $s$,
the corresponding feature-space shift is
\begin{equation}
\Delta q_{kl}(Z_j)=\frac{\Delta v_{kl}(Z_j)}{s}.
\label{eq:supp_dq}
\end{equation}
This derivation supports the main paper's projection and cross-height-shift
equations and clarifies that $P_2$ is a rectified KITTI projection matrix, while
$\mathbf{K}$ denotes the $3\times3$ intrinsic parameters extracted from it. The
same $P_2$ is used consistently for training-time BEV-target generation and for
inference-time sampling.

\section{Deformable Attention over Multi-Height IPM}
\label{sec:bond}
Section~\ref{sec:geom} fixes the projection coordinates by geometry.
The BEV decoder learns to aggregate the resulting height-conditioned evidence.
The attention acts on normalized reference contrasts (Sec.~\ref{sec:ground}),
not directly on raw height maps or a probability distribution over physical height.

\paragraph{The composed operator.}
The raw height-$k$ feature at a BEV grid point $\mathbf{r}=(X,Z)$ is
\begin{equation}
\mathbf{B}_k(\mathbf{r})=
\mathrm{bilinear}\!\left(\mathbf{F}_{\mathrm{fused}},\pi_k(\mathbf{r})\right),
\qquad
\pi_k(\mathbf{r})=
\left(f_x\tfrac{X-t_x}{Z}+c_x,\ f_y\tfrac{Y_k-t_y}{Z}+c_y\right).
\end{equation}
Let $\mathbf{T}_\ell$ denote the ordered, per-level normalized reference and
contrast maps. For one attention head, omitting channel projections for clarity,
\begin{equation}
\mathbf{o}_q=\sum_{\ell=0}^{K-1}\sum_{p=1}^{P} A_{\ell p}\,
\mathrm{bilinear}\!\left(\mathbf{T}_\ell,
\mathbf{r}_q+\Delta\mathbf{r}_{\ell p}\right),\qquad
A_{\ell p}=\mathrm{softmax}_{\ell,p}\big(\theta_{\mathrm{att}}(q)_{\ell p}\big).
\label{eq:bond}
\end{equation}
Thus image-to-plane sampling is followed by reference subtraction,
normalization, and a second interpolation in BEV. Calibration supplies the
geometry-defined evidence, while learned offsets and weights determine how a
query reads it. Because normalization and reference subtraction intervene,
this operator is not generally equivalent to sampling one raw image feature at
a single attention-weighted height.

\paragraph{Height-conditioned evidence.}
At a fixed BEV cell, the raw plane samples share an image column and differ in row
by $\Delta v_{kl}=f_y(Y_k-Y_l)/Z$ (Eq.~\eqref{eq:supp_dv}).
The level arrangement therefore retains a calibrated vertical structure.
After reference subtraction and normalization, attention weights summarize
the usefulness of these evidence levels, not physical-height probabilities.

\paragraph{Intuition.}
Depth-based lifting~\cite{philion2020lift,reading2021caddn} distributes features
along rays according to predicted pixel-depth probabilities. Single-plane IPM
uses one assumed height, whereas MH-IPM supplies several calibrated planes.
Deformable attention~\cite{zhu2021deformable} aggregates the ordered evidence
per query. This avoids predicted pixel depth in the view transform, while
object depth remains predicted by the detector.

\section{Reference-Anchored Representation}
\label{sec:ground}
\begin{figure}[htbp]
\centering
\includegraphics[width=0.98\textwidth]{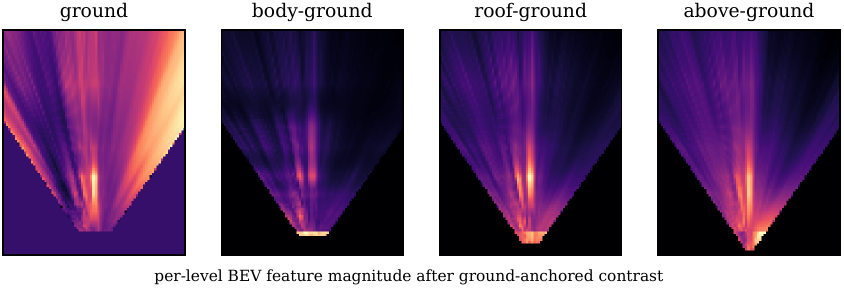}
\caption{\textbf{Per-level BEV feature magnitude after reference-anchored
cross-height contrast (ground in default)} (Eq.~\eqref{eq:supp_anchor}). Panels correspond to the
ground feature $\mathbf{B}'_0$ and the height-minus-ground contrast levels
$\mathbf{B}'_k$ supplied to the BEV decoder. Each panel is normalized independently.}
\label{fig:supp_levels}
\end{figure}

Given the height-plane BEV maps $\{\mathbf{B}_k\}_{k=1}^{K}$ with
$\mathbf{B}_k\in\R^{d\times U_z\times U_x}$, we denote the largest-$Y$ reference
plane by $g$. It coincides with nominal ground, $Y_g=\Hcam$, in the default
configuration. \ours{} reparameterizes the stack as
\begin{equation}
\mathbf{B}'_0=\mathbf{B}_g,
\qquad
\mathbf{B}'_k=\mathbf{B}_k-\mathbf{B}_g,\quad k\neq g.
\label{eq:supp_anchor}
\end{equation}
Before per-level normalization, the transform is invertible:
\begin{equation}
\mathbf{B}_g=\mathbf{B}'_0,
\qquad
\mathbf{B}_k=\mathbf{B}'_k+\mathbf{B}'_0.
\label{eq:supp_anchor_inv}
\end{equation}
Before GroupNorm, Eq.~\eqref{eq:supp_anchor} is an information-preserving linear
reparameterization: the decoder receives the same height layout, expressed as
reference-anchored contrast. For $K=4$ and $Y_1<Y_2<Y_3<Y_4$, the decoder
receives $(\mathbf{B}'_0,\mathbf{B}'_3,\mathbf{B}'_2,\mathbf{B}'_1)$ after per-level
GroupNorm. In the default configuration this emphasizes above-ground deviations
from the ground reference. The test-time interventions in Table~\ref{tab:layout}
show sensitivity to the trained level order. The reference need not coincide exactly with
ground, as the trained coverage variants below demonstrate.
Figure~\ref{fig:supp_levels} visualizes the per-level ground-anchored contrast
maps.

\section{Height-Plane Sensitivity}
\label{sec:plane}
\label{sec:layout}
We probe the number, placement, and ordering of height planes.
Table~\ref{tab:plane} reports height-count and placement results.
The two small placement changes retain the $1.65$\,m reference; they do not
shift every plane together. The random ordered set reaches $25.16$, compared
with $24.83$ for single-plane IPM. These tested sets do not imply that arbitrary
plane placement is harmless.

Table~\ref{tab:layout} instead permutes the trained level order at inference,
without retraining. The large drops indicate learned roles for the ordered
levels, not interchangeable height samples. These inference-only interventions
are distinct from training another model with a different plane set.

The additional trained $K=4$ model with planes at
$\{0.70,1.20,1.70,2.10\}$\,m reaches $25.30$ Moderate $\APThreeD$.
The set $\{-0.20,0.30,0.80,1.20\}$\,m reaches $24.33$.
Both results are three-seed means under the full training recipe. The sets span $1.40$\,m at different placements
and omit nominal ground at $1.65$\,m. Every model retains the largest-$Y$
reference and ordered contrast construction. These results support suitable
vertical coverage rather than a mandatory exact ground plane, without
establishing a unique optimal placement.

\begin{table}[htbp]
\centering
\caption{\textbf{Height-plane sensitivity of trained models} (KITTI \textit{val}, Car, Moderate).
\emph{Top}: number of planes $K$ (context; matches the main-paper $K$-ablation).
\emph{Bottom}: two placement changes retaining the $1.65$\,m reference,
and one random ordered set. The two small changes stay within $0.1$ AP of the
default, shown in \textbf{bold}.}
\label{tab:plane}
\footnotesize
\begin{tabular}{llcc}
\toprule
Height set (m) & Interpretation & $\APThreeD$ Mod. & $\APBEV$ Mod. \\
\midrule
$\{1.65\}$ & single ground plane & 24.83 & 31.18 \\ $\{0.25,1.65\}$ & top + ground & 25.34 & 31.92 \\
$\{0.25,1.0,1.65\}$ & coarse 3-plane & 25.64 & 32.36 \\
$\{0.25,0.75,1.25,1.65\}$ & default ($K{=}4$) & \textbf{25.84} & \textbf{32.66} \\ \midrule
$\{0.45,0.85,1.25,1.65\}$ & lower range & 25.77 & 32.52 \\ $\{0.10,0.60,1.10,1.65\}$ & higher range & 25.80 & 32.57 \\ $\{-0.85,0.40,1.10,2.05\}$ & random, ordered & 25.16 & 31.13 \\ \bottomrule
\end{tabular}
\vspace{6pt}
\end{table}

\begin{table}[htbp]
\centering
\caption{\textbf{Height-layout robustness} (KITTI \textit{val}, Car,
$\APThreeD$ Mod.). Inference-only permutations of the trained level order;
the checkpoint and weights are unchanged.}
\label{tab:layout}
\footnotesize
\begin{tabular}{llcc}
\toprule
Intervention & Type & $\APThreeD$ Mod. & $\Delta$ \\
\midrule
Full \ours{} (ordered) & --- & \textbf{25.84} & --- \\
Test-time shuffle (mean of 4) & inference & $14.35${\scriptsize$\,\pm3.9$} & $-11.49$ \\ Test-time reverse & inference & 11.17 & $-14.67$ \\ \bottomrule
\end{tabular}
\end{table}
\FloatBarrier

\section{Training Stability}
\label{sec:seed}
Across ten independent runs of the full $K{=}4$ model with the fixed recipe,
Car Moderate $\APThreeD$ on KITTI \textit{val} has mean $25.27$, standard
deviation $0.30$, and range $24.80$--$25.84$ (Table~\ref{tab:seed}).
The reported score is the best run.

\begin{table}[htbp]
\centering
\caption{\textbf{Training stability of \ours{} across 10 seeds on KITTI \textit{val}}
(Car, $\APThreeD$ Moderate, IoU~$0.7$). Min, max, and mean\,$\pm$\,std over ten
independent training runs of the full $K{=}4$ model; the best run corresponds to our
reported result.}
\label{tab:seed}
\footnotesize
\begin{tabular}{lccc}
\toprule
Method & Min & Max & Mean $\pm$ Std \\
\midrule
\ours{} ($K{=}4$, full) & 24.80 & \textbf{25.84} & $25.27 \pm 0.30$ \\
\bottomrule
\end{tabular}
\end{table}

\section{Occlusion and Truncation Breakdowns}
\label{sec:break}
Tables~\ref{tab:occ}
and~\ref{tab:trunc} stratify by occlusion and truncation: for each level we keep
only that level's Car objects as evaluation targets (other levels become ignore
regions) and report Car $\APThreeD$ (R40) at the Hard difficulty setting, which
admits every level. \ours{} leads at all three occlusion levels, and its
relative margin grows with occlusion, more than doubling the MonoDGP score on
largely-occluded objects ($1.49$ vs.\ $0.65$), consistent with the BEV
representation helping most where perspective depth is least observable.

\begin{table}[htbp]
\centering
\begin{minipage}[t]{0.48\textwidth}
\centering
\caption{\textbf{Occlusion breakdown} (Car $\APThreeD$, Hard).}
\label{tab:occ}
\footnotesize
\begin{tabular}{lccc}
\toprule
Method & Full & Partly & Largely \\
\midrule
MonoDGP & 17.14 & 5.76 & 0.65 \\
MonoCoP & 16.88 & 6.29 & 0.86 \\
\ours{}  & \textbf{18.23} & \textbf{7.40} & \textbf{1.49} \\
\bottomrule
\end{tabular}
\end{minipage}
\hfill
\begin{minipage}[t]{0.48\textwidth}
\centering
\caption{\textbf{Truncation breakdown} (Car $\APThreeD$, Hard).}
\label{tab:trunc}
\footnotesize
\begin{tabular}{lccc}
\toprule
Method & Low & Med. & High \\
\midrule
MonoDGP & 16.72 & \textbf{2.85} & 1.03 \\
MonoCoP & 18.17 & 1.94 & 0.78 \\
\ours{}  & \textbf{20.84} & 2.58 & \textbf{1.18} \\
\bottomrule
\end{tabular}
\end{minipage}
\end{table}

\section{Results on the Waymo Open Dataset}
\label{sec:waymo}
We additionally benchmark \ours{} on the Waymo Open Dataset~\cite{sun2020scalability},
following the DEVIANT~\cite{kumar2022deviant}\,/\,MonoDGP~\cite{pu2025monodgp}
front-camera protocol ($52{,}386$ train / $39{,}848$ val, Vehicle class, official
$\mathrm{AP_{3D}}$\,/\,$\mathrm{APH_{3D}}$), with the KITTI-default height planes
unchanged. Table~\ref{tab:waymo} reports the results.

\begin{table}[htbp]
\centering
\caption{\textbf{Waymo Open Dataset \textit{val} Vehicle results.} All methods are
trained on the Waymo front-camera split (DEVIANT~\cite{kumar2022deviant} protocol;
$52{,}386$ train / $39{,}848$ val) and scored with the official Waymo
$\mathrm{AP_{3D}}$ / heading-weighted $\mathrm{APH_{3D}}$ at Level~1 / Level~2 and
IoU$_\mathrm{3D}\!\in\!\{0.5,0.7\}$, broken down by range (All / $[0,30)$ / $[30,50)$ /
$[50,\infty)$\,m). \ours{} uses the KITTI-default metric height planes and leads on the
mid-far-range bins at both thresholds. [Key: \textbf{first}, \underline{second};
$^\dagger$: retrained by~\cite{zhang2026monocop}.]}
\label{tab:waymo}
\small
\setlength{\tabcolsep}{2.5pt}
\begin{tabular*}{\textwidth}{@{\extracolsep{\fill}}llrrrrrrrr@{}}
\toprule
\multicolumn{10}{c}{\textbf{(a) IoU$_\mathrm{3D}\geq0.5$}} \\
\midrule
Method & Diff. & \multicolumn{4}{c}{$\mathrm{APH_{3D}}$} & \multicolumn{4}{c}{$\mathrm{AP_{3D}}$} \\
\cmidrule(lr){3-6}\cmidrule(lr){7-10}
 & & All & 0--30 & 30--50 & 50--$\infty$ & All & 0--30 & 30--50 & 50--$\infty$ \\
\midrule
GUP Net~\cite{lu2021gupnet} & \multirow{7}{*}{L1} & 9.94 & 24.59 & 4.78 & 0.22 & 10.02 & 24.78 & 4.84 & 0.22 \\
DEVIANT~\cite{kumar2022deviant} &  & 10.89 & \underline{26.64} & 5.08 & 0.18 & 10.98 & \underline{26.85} & 5.13 & 0.18 \\
MonoDETR$^\dagger$~\cite{zhang2023monodetr} &  & 9.60 & 23.58 & 4.67 & 0.99 & 9.68 & 23.78 & 4.72 & 1.00 \\
MonoUNI~\cite{jinrang2023monouni} &  & 10.73 & 26.30 & 3.98 & 0.55 & 10.98 & 26.63 & 4.04 & 0.57 \\
MonoDGP$^\dagger$~\cite{pu2025monodgp} &  & 9.84 & 23.73 & 5.01 & 0.98 & 10.06 & 24.01 & 5.06 & 0.99 \\
MonoCoP~\cite{zhang2026monocop} &  & \underline{11.65} & \textbf{27.35} & \underline{5.97} & \underline{1.46} & \underline{11.76} & \textbf{27.59} & \underline{6.03} & \underline{1.48} \\
\textbf{\ours{} (Ours)} &  & \textbf{11.89} & 24.31 & \textbf{9.02} & \textbf{2.21} & \textbf{12.05} & 24.59 & \textbf{9.18} & \textbf{2.25} \\
\midrule
GUP Net~\cite{lu2021gupnet} & \multirow{7}{*}{L2} & 9.31 & 24.50 & 4.62 & 0.19 & 9.39 & 24.69 & 4.67 & 0.19 \\
DEVIANT~\cite{kumar2022deviant} &  & 10.20 & \underline{26.54} & 4.90 & 0.16 & 10.29 & \underline{26.75} & 4.95 & 0.16 \\
MonoDETR$^\dagger$~\cite{zhang2023monodetr} &  & 9.00 & 23.49 & 4.51 & 0.86 & 9.08 & 23.70 & 4.55 & 0.87 \\
MonoUNI~\cite{jinrang2023monouni} &  & 10.24 & 26.24 & 3.89 & 0.51 & 10.38 & 26.57 & 3.95 & 0.53 \\
MonoDGP$^\dagger$~\cite{pu2025monodgp} &  & 9.32 & 23.65 & 4.84 & 0.85 & 9.43 & 23.92 & 4.88 & 0.86 \\
MonoCoP~\cite{zhang2026monocop} &  & \underline{10.93} & \textbf{27.25} & \underline{5.76} & \underline{1.27} & \underline{11.03} & \textbf{27.49} & \underline{5.82} & \underline{1.29} \\
\textbf{\ours{} (Ours)} &  & \textbf{11.12} & 24.25 & \textbf{8.68} & \textbf{1.91} & \textbf{11.27} & 24.52 & \textbf{8.83} & \textbf{1.94} \\
\bottomrule
\end{tabular*}

\medskip
\begin{tabular*}{\textwidth}{@{\extracolsep{\fill}}llrrrrrrrr@{}}
\toprule
\multicolumn{10}{c}{\textbf{(b) IoU$_\mathrm{3D}\geq0.7$}} \\
\midrule
Method & Diff. & \multicolumn{4}{c}{$\mathrm{APH_{3D}}$} & \multicolumn{4}{c}{$\mathrm{AP_{3D}}$} \\
\cmidrule(lr){3-6}\cmidrule(lr){7-10}
 & & All & 0--30 & 30--50 & 50--$\infty$ & All & 0--30 & 30--50 & 50--$\infty$ \\
\midrule
GUP Net~\cite{lu2021gupnet} & \multirow{7}{*}{L1} & 2.27 & 6.11 & 0.80 & 0.03 & 2.28 & 6.15 & 0.81 & 0.03 \\
DEVIANT~\cite{kumar2022deviant} &  & 2.67 & 6.90 & 0.98 & 0.02 & 2.69 & 6.95 & 0.99 & 0.02 \\
MonoDETR$^\dagger$~\cite{zhang2023monodetr} &  & 2.10 & 5.94 & 0.73 & 0.12 & 2.11 & 5.99 & 0.73 & 0.12 \\
MonoUNI~\cite{jinrang2023monouni} &  & \textbf{3.16} & \textbf{8.50} & 0.86 & 0.12 & \textbf{3.20} & \textbf{8.61} & 0.87 & 0.13 \\
MonoDGP$^\dagger$~\cite{pu2025monodgp} &  & 2.39 & 6.62 & 0.84 & 0.12 & 2.41 & 6.67 & 0.84 & 0.12 \\
MonoCoP~\cite{zhang2026monocop} &  & \underline{2.70} & \underline{7.38} & \underline{1.06} & \underline{0.16} & \underline{2.72} & \underline{7.44} & \underline{1.07} & \underline{0.16} \\
\textbf{\ours{} (Ours)} &  & 2.59 & 5.86 & \textbf{1.70} & \textbf{0.41} & 2.62 & 5.91 & \textbf{1.72} & \textbf{0.41} \\
\midrule
GUP Net~\cite{lu2021gupnet} & \multirow{7}{*}{L2} & 2.12 & 6.08 & 0.77 & 0.02 & 2.14 & 6.13 & 0.78 & 0.02 \\
DEVIANT~\cite{kumar2022deviant} &  & 2.50 & 6.87 & 0.94 & 0.02 & 2.52 & 6.93 & 0.95 & 0.02 \\
MonoDETR$^\dagger$~\cite{zhang2023monodetr} &  & 1.97 & 5.92 & 0.70 & 0.10 & 1.98 & 5.96 & 0.71 & 0.10 \\
MonoUNI~\cite{jinrang2023monouni} &  & \textbf{3.00} & \textbf{8.48} & 0.84 & 0.12 & \textbf{3.04} & \textbf{8.59} & 0.85 & 0.12 \\
MonoDGP$^\dagger$~\cite{pu2025monodgp} &  & 2.24 & 6.59 & 0.81 & 0.10 & 2.26 & 6.65 & 0.81 & 0.10 \\
MonoCoP~\cite{zhang2026monocop} &  & \underline{2.53} & \underline{7.35} & \underline{1.02} & \underline{0.14} & \underline{2.55} & \underline{7.41} & \underline{1.03} & \underline{0.14} \\
\textbf{\ours{} (Ours)} &  & 2.42 & 5.84 & \textbf{1.63} & \textbf{0.35} & 2.45 & 5.89 & \textbf{1.66} & \textbf{0.35} \\
\bottomrule
\end{tabular*}
\end{table}

\begin{figure}[p]
\centering
\includegraphics[height=0.82\textheight]{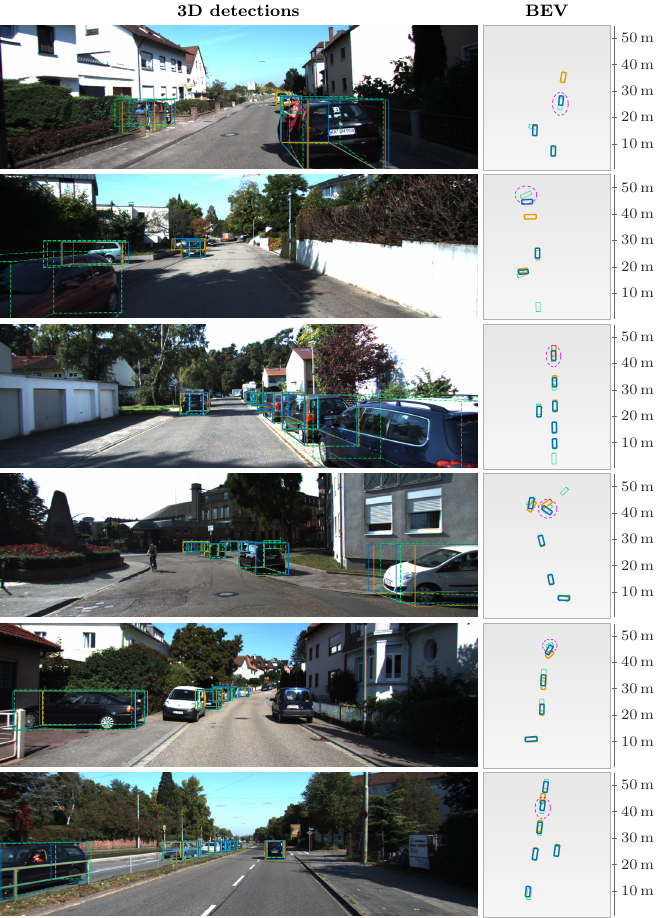}
\caption{\textbf{Qualitative KITTI \textit{val} comparison: \ours{} vs.\ MonoCoP.}
Six validation scenes (rows), each with the image-space 3D-detection panel
(\emph{left}) and the metric BEV panel (\emph{right}, depth ticks: $10$--$50$\,m). Green dashed
boxes are ground truth, \textcolor{blue}{blue} boxes are \ours{}, and
\textcolor{orange}{orange} boxes are MonoCoP; dashed magenta circles mark telling
objects. \ours{}'s blue boxes consistently sit closer to the green ground truth in
the BEV panel than MonoCoP's orange. This is clearest on the highlighted mid- and
far-range vehicles, where MonoCoP misplaces depth along the viewing ray while
\ours{} keeps the metric footprint aligned.}
\label{fig:supp_grid}
\end{figure}

\section{Qualitative Results}
\label{sec:qual}
Figure~\ref{fig:supp_grid} compares \ours{} against MonoCoP~\cite{zhang2026monocop}
on six KITTI \textit{val} scenes. \ours{} consistently places vehicles closer to
their true metric BEV footprint, most visibly at mid and far range where
single-image depth is least reliable and small BEV-alignment errors dominate the
3D IoU.

\begin{figure}[htbp]
\centering
\includegraphics[width=\textwidth]{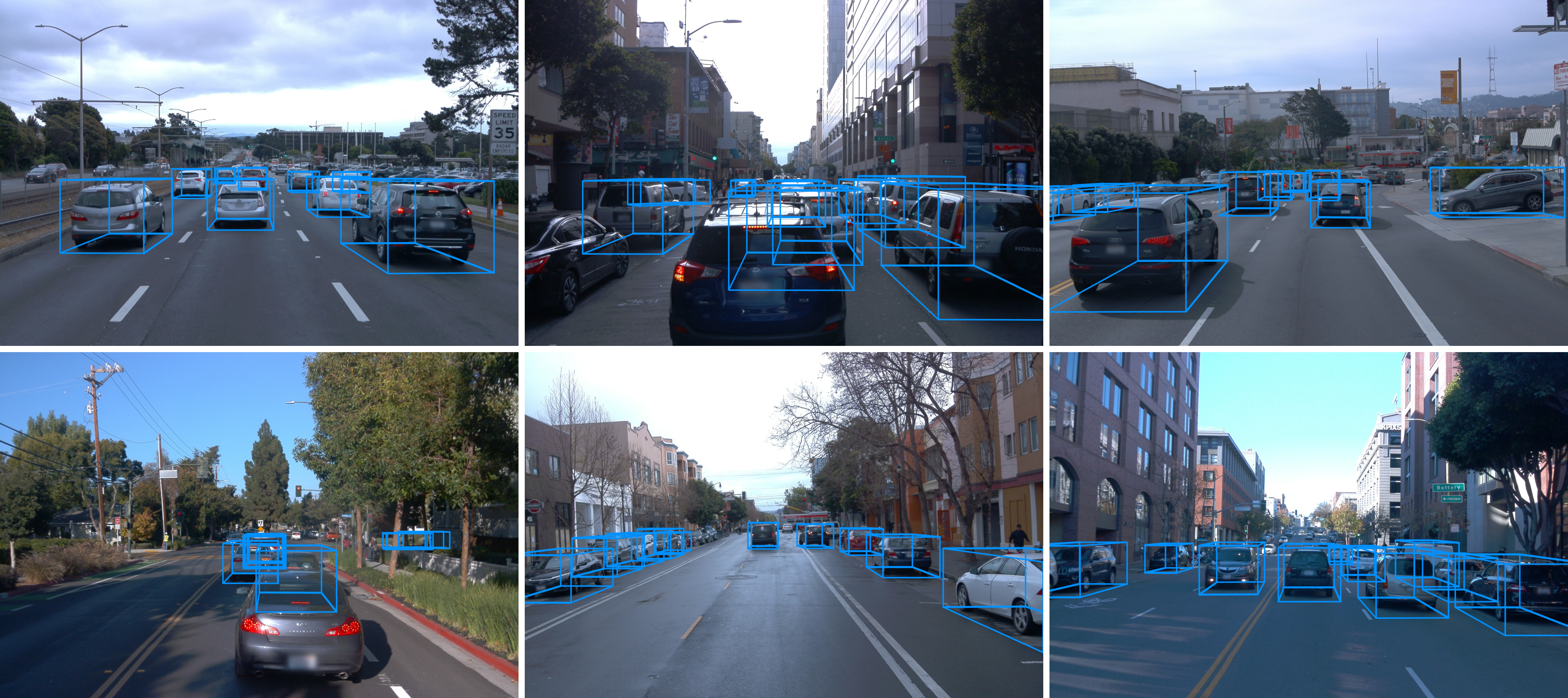}
\caption{\textbf{Qualitative \ours{} detections on the Waymo Open Dataset}
(front camera, \textit{val}). Predicted 3D boxes (\textcolor{blue}{blue}) from our
reported model on six validation scenes. \ours{} localizes dense traffic across
highway and urban scenes under the KITTI-default metric height planes.}
\label{fig:supp_waymo_qual}
\end{figure}

Figure~\ref{fig:supp_waymo_qual} shows \ours{}'s predicted 3D boxes on the Waymo
Open Dataset across six validation scenes.

\begin{figure}[htbp]
\centering
\begin{minipage}{0.49\textwidth}\centering
\includegraphics[width=\textwidth]{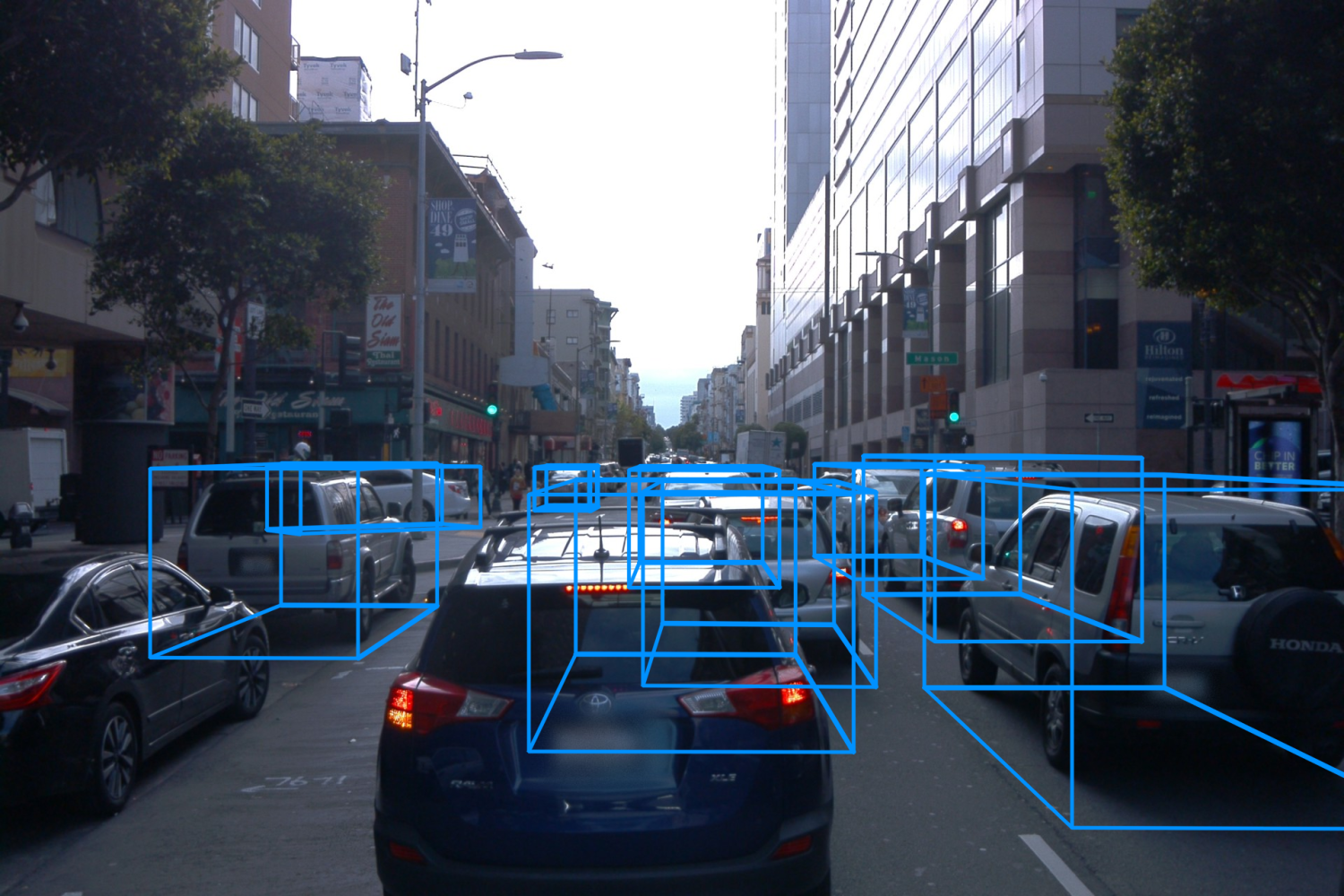}\\[3pt]
{\footnotesize (a) KITTI-default planes}
\end{minipage}\hfill
\begin{minipage}{0.49\textwidth}\centering
\includegraphics[width=\textwidth]{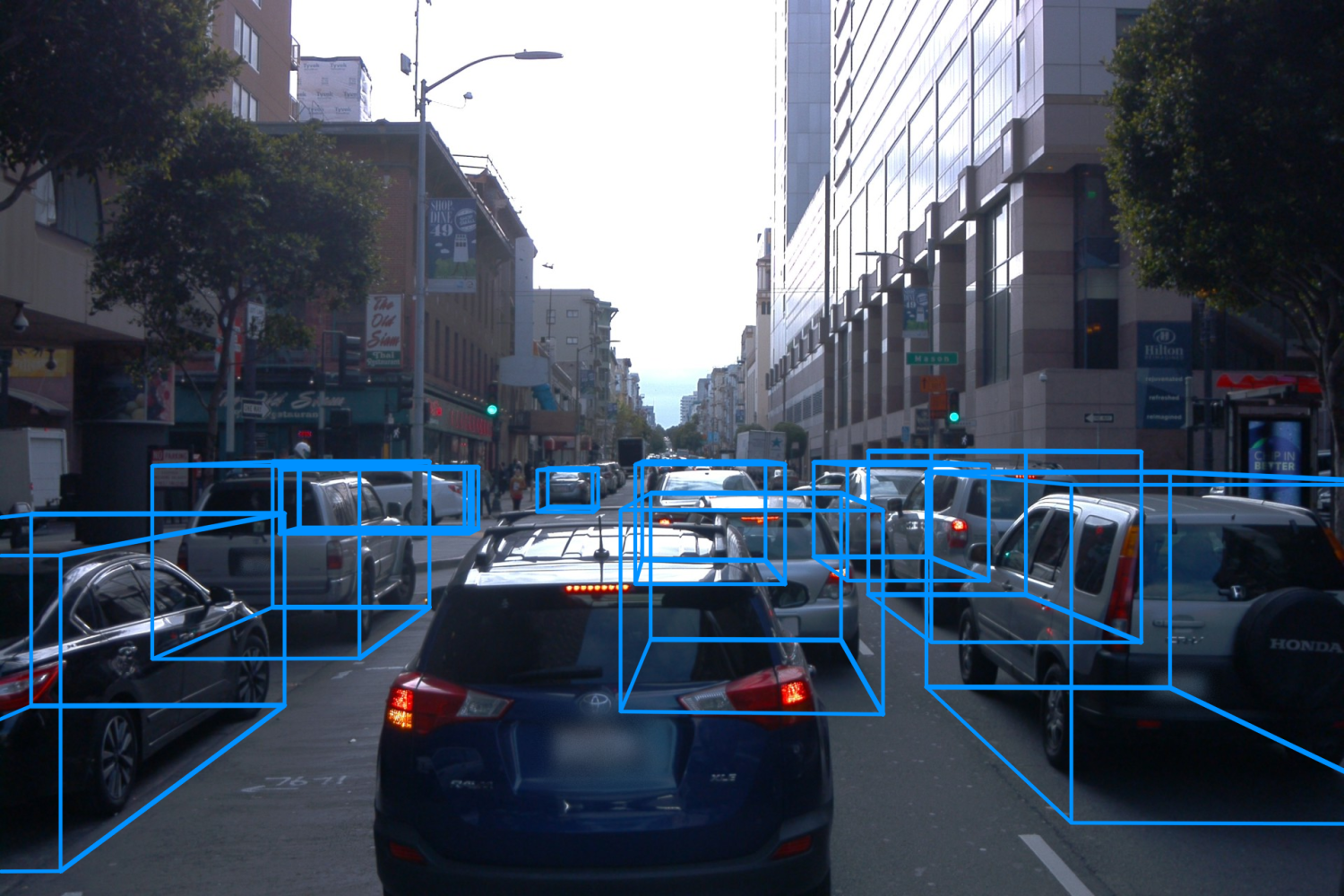}\\[3pt]
{\footnotesize (b) Shifted-plane model}
\end{minipage}
\caption{\textbf{Different plane placements on a Waymo scene.} Predicted 3D
boxes (\textcolor{blue}{blue}) from (a) the KITTI-default and (b) shifted-plane
models. The bottom-left vehicle is missed in (a) but detected in (b).
This qualitative example motivates examining vertical coverage, but different
trained checkpoints prevent attributing the detection change to planes alone.}
\label{fig:supp_waymo_ipm_failure}
\end{figure}

Figure~\ref{fig:supp_waymo_ipm_failure} compares models with different plane
placements on one Waymo scene. A nearby vehicle missed by the KITTI-default
model is detected by the shifted-plane model. The checkpoints differ, so this
is an illustrative comparison, not an isolated plane-placement ablation.

\section{Failure Cases and Limitations}
\label{sec:fail}
\noindent\textbf{Height coverage.}
The plane range is shared across categories, but a finite set does not cover
arbitrary scene geometry. Section~\ref{sec:plane} distinguishes the
plane-placement results, inference-only order changes, and trained
coverage variants. The latter omit an exact ground plane yet retain useful
performance. Their different placements and equal spans show why neither
ground coincidence nor span width alone explains accuracy. The reported
Waymo benchmark uses unchanged KITTI-default planes; the qualitative shifted-plane
comparison does not establish calibration-free transfer. Scene-adaptive plane
selection remains future work.

\medskip\noindent\textbf{Calibration and camera height.}
\ours{} assumes known calibration and a roughly fixed camera height.
Tables~\ref{tab:calib} and~\ref{tab:attribution} examine complementary sources
of calibration sensitivity. Table~\ref{tab:calib} reports the camera-height
study: all IPM planes are shifted together, and a $\pm0.10$\,m offset costs
roughly $2$--$3$ AP. This is a limitation of the assumed height in MH-IPM.

Table~\ref{tab:attribution} reports the focal-sensitivity results and their component
attribution. The full-network rows scale both $f_x$ and $f_y$ in the network
input for each detector. The component rows change only the geometric-depth
$f_x$ term or the MH-IPM $(f_x,f_y)$ sampling coordinates, keeping all other
calibration uses unchanged. The larger effect comes through geometric depth,
consistent with its inherited focal dependence. These component effects
should not be summed or equated with the full-network intervention.
Images and ground truth stay unchanged, and final coordinate decoding retains
the original calibration. This is a metadata-error test, not a pitch or
sloped-road evaluation.

The focal-sensitivity experiments use the default inference setting in Sec.~\ref{sec:impl}.

\begin{table}[htbp]
\centering
\caption{\textbf{Camera-height sensitivity under inference-time perturbations}
(KITTI \textit{val}, Car, $\APThreeD$ Mod.), measured on the \ours{} checkpoint.
All IPM planes are offset together without retraining. Focal perturbations
are reported separately in Table~\ref{tab:attribution}.}
\label{tab:calib}
\footnotesize
\begin{tabular}{lcc}
\toprule
Perturbation & $\APThreeD$ Mod. & $\Delta$ \\
\midrule
none & \textbf{25.84} & --- \\ $\Hcam-0.20$ m & 21.98 & $-3.86$ \\
$\Hcam-0.10$ m & 23.54 & $-2.30$ \\
$\Hcam+0.10$ m & 22.66 & $-3.17$ \\
$\Hcam+0.20$ m & 19.21 & $-6.63$ \\
\bottomrule
\end{tabular}
\end{table}

\begin{table}[htbp]
\centering
\caption{\textbf{Focal-sensitivity attribution.} Absolute Car Moderate $\APThreeD$,
inference only. Top: both network-input focal axes
perturbed. Bottom: only the geometric-depth $f_x$ term or MH-IPM $(f_x,f_y)$
perturbed.}
\label{tab:attribution}
\small
\begin{tabular}{lrr}
\toprule
Model / intervention & $-2.5\%$ AP & $+2.5\%$ AP \\
\midrule
MonoDGP, full network & 15.82 & 14.25 \\
MonoCoP, full network & 15.92 & 14.73 \\
\ours{}, full network & 15.42 & 14.24 \\
\midrule
\ours{}, geometric depth only & 17.32 & 17.20 \\
\ours{}, MH-IPM only & 24.37 & 23.62 \\
\bottomrule
\end{tabular}
\end{table}
The results in Table~\ref{tab:attribution} use the calibration-sweep setting
and do not replace the published benchmark results.

\noindent\textbf{Range limits.} Two geometric limits remain. At far range the
cross-height row separation shrinks toward the feature stride, so the planes become
hard to distinguish and the BEV cue weakens, consistent with the narrowing
$40$--$60$\,m gains in the main paper. At the strict IoU$_\mathrm{3D}\!\geq\!0.7$
operating point on Waymo, the residual gap concentrates in the \emph{near} range, where
vehicles are large and frequently truncated at the image border, so small height,
dimension, or truncation errors incur a disproportionate IoU penalty; we verified this
is geometric rather than a ranking or post-processing artifact (sweeping top-$k$,
toggling BEV NMS, altering the score, and filtering tiny boxes all leave it unchanged).

\medskip\noindent\textbf{Small and hard classes.}
The coarse $96{\times}80$ grid can limit spatial detail for thin objects, but
resolution is not a complete explanation of small-class errors. The finer-grid
experiment in Sec.~\ref{sec:additional} improves Cyclist more than Pedestrian,
at greater latency and memory cost. MonoDGP~\cite{pu2025monodgp} also discusses
irregular Cyclist shape and greater projection uncertainty. That is a plausible
factor, not a cause isolated by our resolution ablation.

\section{Additional Ablations}
\label{sec:additional}
This section reports additional ablation studies.
Detection scores are KITTI \textit{val} Moderate R40 AP, at IoU $0.7$ for Car
and $0.5$ for Pedestrian/Cyclist. Additional trained variants use the full
250-epoch recipe and report three-seed means. The reported $25.84$ reference
is the best of ten runs, whose distribution is reported in Sec.~\ref{sec:seed}.

\paragraph{Decoder composition.}
The 2D-only stage predicts class and box fields, without depth cross-attention.
The 3D stage is our depth-guided image decoder, and the BEV stage
uses the same metric refinement as the full model. Adding a 2D stage before
3D+BEV increases parameters and latency without an observed accuracy gain.
The $24.99$ result is close to the full model's ten-seed mean, so it does not
establish an accuracy difference beyond seed variation. The 2D+BEV variant
initializes metric coordinates from projected features without perspective
depth and size anchors. Its lower result supports the value of these anchors,
but does not rule out other 2D+BEV designs. Published MonoDGP supplies the
2D+3D reference in the main benchmark tables; it is not a controlled retraining.

\begin{table}[htbp]
\centering
\caption{\textbf{Decoder and view-transform ablations.} Car Moderate
$\APThreeD$. $^*$: three-seed mean; $^\dagger$: reported model
reference (Sec.~\ref{sec:seed}). Costs are from a separate RTX~3090 profiling
session, batch 1, FP32, forward only. Memory is peak allocated VRAM.}
\label{tab:additional}
\small
\begin{tabular}{lrrrr}
\toprule
Variant & 3D AP & Params (M) & ms/image & MiB \\
\midrule
2D + BEV$^*$ & 21.02 & 40.389 & 26.939 & 459.269 \\
3D + BEV$^\dagger$ & 25.84 & 39.856 & 28.137 & 457.230 \\
2D + 3D + BEV$^*$ & 24.99 & 41.736 & 29.174 & 464.403 \\
\midrule
3D + BEV, LSS$^*$ & 24.05 & 39.856 & 30.545 & 457.230 \\
\bottomrule
\end{tabular}
\end{table}

\paragraph{Controlled depth lifting.}
The LSS-style replacement keeps the fused features, predicted 80-bin depth
distribution, BEV range and $96\times80$ grid, queries, decoder, losses,
matching, augmentation, and schedule fixed. Its three-seed mean is $24.05$ AP
and maximum $24.17$, below all ten MH-IPM runs ($24.80$--$25.84$).
MH-IPM has $7.9\%$ lower measured latency at the same parameter count and
allocated memory. This supports the IPM representation within the hybrid
pathway, rather than attributing the entire gain to an added decoder.
The seed comparison is descriptive, not a paired statistical test.

\paragraph{Resolution and occupancy.}
The $192\times160$ study retains the metric BEV range and updates target
rasterization and spatial shapes to the finer grid. Cyclist Moderate 3D AP
rises from $8.32$ to $9.11$ ($+0.79$), while Pedestrian changes by only $+0.12$.
Latency rises from $28.137$ to $36.697$\,ms and allocated memory from
$457.230$ to $887.305$\,MiB. Thus resolution helps Cyclist at a substantial
cost, but does not resolve all small-object errors.

Removing $\mathcal L_{\rm bmap}$ gives Car 3D AP $25.75$ ($-0.09$) and
BEV AP $32.07$ ($-0.59$). This is consistent with a modest training-only
role in reinforcing footprint-aligned features, rather than a large 3D AP
contribution. The occupancy head is unused at inference.

\paragraph{Orientation.}
Final predictions attain $88.89$ Moderate $\mathrm{AOS_{R40}}$.
Their heading is predicted from BEV query features with per-layer bin/residual
supervision, not copied from the image branch. This is a final-detection
orientation score, not an isolated angular-error or matched-branch comparison.

\FloatBarrier

\bibliographystyle{splncs04}
\bibliography{supplement_references}